\pdfoutput=1
\documentclass[11pt]{article}

\usepackage[final]{acl}

\usepackage{times}
\usepackage{latexsym}

\usepackage[T1]{fontenc}
\usepackage[utf8]{inputenc}

\usepackage{microtype}

\usepackage{inconsolata}

\usepackage{graphicx}

\usepackage{amssymb}
\usepackage{multirow}
\usepackage{multicol}
\usepackage{makecell}
\usepackage{amsmath}
\usepackage[table]{xcolor}
\usepackage{arydshln}
\usepackage{tikz}

\makeatletter
\def\dottedline{%
  \noalign{\global\arrayrulewidth=0.3pt} % Adjust dot thickness
  \hdashline[0.5pt/2pt]  % Adjust dot spacing (dot size / gap size)
  \noalign{\global\arrayrulewidth=0.4pt} % Reset line thickness
}
\makeatother

\title{Multi-Task Composability of QLoRA PEFT Modules}
\title{QLoRA PEFT Modules Can be Composed for Multi-Task Capabilities}
\title{Composability of QLoRA PEFT Modules for Multi-Task Capabilities}
\title{Composability of QLoRA PEFT Modules in Plug-and-Play Scenarios}
\title{Output Composability of QLoRA PEFT Modules for Plug-and-Play Attribute-Controlled Text Generation}

\author{Michela Lorandi \and Anya Belz \\
  ADAPT, Dublin City University \\
  Dublin, Ireland \\
  \texttt{\{michela.lorandi2,anya.belz\}@mail.dcu.ie} }

\begin{document}
\maketitle
\begin{abstract}
Parameter-efficient fine-tuning (PEFT) techniques offer task-specific fine-tuning at a fraction of the cost of full fine-tuning, but require separate fine-tuning for every new task (combination). In this paper, we explore three ways of generalising beyond single-task training/inference: (i) training on combinations of multiple, related datasets; (ii) at inference, composing the \textit{weight matrices} of separately trained PEFT modules; and (iii) at inference, composing the \textit{outputs} of separately trained PEFT modules. We test these approaches on three different LLMs, QLoRA as the PEFT technique, and three sets of controlled text generation datasets for sentiment control, topic control, and multi-attribute control. We find that summing PEFT module outputs is a particularly strong composition method, which consistently either outperforms or matches the performance of alternative approaches. This is the case even when comparing against single-task specialised modules on the single-task test set, where three-module output composition achieves an average 2\% point performance \textit{increase} across all models for sentiment control. 
\end{abstract}

\section{Introduction}
\label{sec:intro}

Given the substantial costs of training state-of-the-art language models, parameter-efficient finetuning (PEFT) techniques such as Adapters \cite{pmlr-v97-houlsby19a}, Prefix Tuning \cite{li-liang-2021-prefix}, and LoRA \cite{hu2021lora} are have become an important part of the toolboxfor task-specific adaptation of pretrained models. PEFT techniques produce parameter matrices that are modular in the sense that they can be attached, detached and replaced in the same host model as needed, and this holds out the appealing (but currently far from realisable) vision of full plug-and-playability where individual task-specifically trained modules can be combined at will to achieve specific model behaviours, potentially even in conjunction with different host models \cite{sabry-belz-2024-assessing}. 

One aim in combining modules in this way is to achieve the combined functionality, e.g.\ controlling multiple attributes of generated text for which the individual modules have been (separately) trained. However, it is also desirable to preserve performance at the separate tasks. 
We examine the extent to which (i) composite-task performance can be achieved, and (ii) single-task performance preserved, when QLoRA PEFT modules are combined in host models. We test the \textbf{composability} of PEFT modules in this sense with different composition techniques on the composite task of multiple-attribute control, and the single tasks of sentiment and topic control. We compare output and parameter composition, as well as a baseline of training a single module on multiple datasets.

We focus on attribute-controlled text generation tasks, because these well established tasks enable systematic evaluation across multiple datasets while providing a controlled setting to isolate the core mechanics of PEFT module composition. 

In$\:$presentation$\:$order,$\:$our$\:$main$\:$contributions$\:$are:

\vspace{-.2cm}
\begin{enumerate}\itemsep=-3pt
    \item A new plug-and-play implementation of QLoRA supporting PEFT module composition for any number of modules, with three different composition techniques (Section~\ref{sec:plug-play-qlora});
    \item New output composition techniques for PEFT modules (Section~\ref{sec:composition-techniques});
    \item A new method for combining disparate datasets into a single, representative dataset (Section~\ref{sec:sampling}); and
    \item New test results for all composition techniques on 3 models, 14 datasets, and 3 composition techniques (Section~\ref{sec:res}). 
\end{enumerate}

\section{Related Research}
\label{sec:rel_research}
\vspace{-.15cm}
Parameter-efficient fine-tuning (PEFT) has been shown \citep{liu2024gpt, hu2021lora,poth2023adapters,whitehouse-etal-2024-low} to effectively inject task-specific knowledge into pretrained models, including in the case of raw models \citep{zhao2024lora} where (most of) the task-level knowledge must be acquired during finetuning. 

PEFT modules have been tested for cross-task transferability,  with prompt tuning \cite{su-etal-2022-transferability,vu-etal-2022-spot}, and other PEFT techniques, where e.g.\ \citet{Ding2023} showed that PEFT-tuning, e.g.\ with LoRA, maintains performance on other tasks only when they are closely related. 

Multiple LoRA modules have been used in mixture-of-expert set-ups, where the single most suitable module is selected on the fly for a given task \citep{feng-etal-2024-mixture,dou-etal-2024-loramoe}. Recent work has also explored rank-wise clustering approaches to LoRA merging \citep{zhao2025merging}. Most similar to our work, two papers have tested combining LoRA modules trained on general language tasks by computing the weighted sums of their parameters (with weights hardwired or optimised) \cite{asadi2024combining}; or by (gradient-free) combinatorial optimisation for automatic selection of modules \cite{huang2023lorahub}. The downsides are (i) additional learning steps after composition, and (ii) inefficient weight learning across many modules despite limited supervised data.

Our work differs from  \citet{asadi2024combining} in three key ways: (1) we introduce output composition methods which to our knowledge have not been previously explored for PEFT modules; (2) we focus on true plug-and-play composition without any post-composition training or optimization; and (3) we demonstrate that output composition not only generalises across related tasks but can even improve individual module performance on their original tasks.

The scenario we address is more ambitious: we want to take multiple task-trained LoRA modules off the shelf, combine them without further learning steps in a host model, to achieve performant results on each of the tasks for which we have loaded a module, as well as on combined tasks requiring their combined functionality. To the best of our knowledge, composability in this sense has not so far been explored; demonstrating it for QLoRA modules for the first time is our aim in this paper. 

Our output composition techniques can be understood through the lens of representation engineering and activation steering \citep{turner2023activation,zou2023representation}, where model behaviour is controlled by manipulating internal representations. When we sum PEFT module outputs, we in effect combine learned steering directions in the model's representation space. Unlike typical activation steering approaches which require manual direction finding through techniques like contrastive activation addition or additional optimization steps \citep{li2023inference}, our approach leverages task-specific PEFT modules as pre-learned steering vectors that can be composed without further training. This connection helps explain why output summing proves particularly effective: each PEFT module captures a task-specific representational direction, and their linear combination naturally integrates these learned behaviours.

\section{Plug-and-play QLoRA}
\label{sec:plug-play-qlora}

We reimplemented\footnote{Available from \url{https://github.com/anonymised}} QLoRA with additional functionality to support the attachment of multiple PEFT modules, and their composition with different  techniques. In vanilla single-module QLoRA, a PEFT block consisting of a down-projection, and up-projection is attached (as shown in Figure~\ref{fig:qlora} in Appendix) in parallel to every key, query, value and feed-forward layer in every transformer block. 

Adopting the notation from \citet{asadi2024combining}, QLoRA learns a far smaller set of task-specific parameters $\Delta\Theta$ in conjunction with the frozen pre-trained model $\Theta_0$. $\Delta\Theta$ (the QLoRA module) consists of trainable low-rank decomposition matrices added as adapter modules to the query, key, value and feed-forward weight matrices in the frozen model $\Theta_0$. In other words, QLoRA adds task-specific parameters $\mathbf{\Delta W}$ to pre-trained frozen weight matrices $\mathbf{W}_0 \in \mathbb{R}^{d\times c}$ in $\Theta_0$:

\vspace{-.3cm}
\[\mathbf{\hat{W}} = \mathbf{W_0} + \frac{\alpha}{r} \mathbf{\Delta W}\]
\vspace{-.5cm}

\noindent where $\mathbf{\Delta W} = \mathbf{AB^T}$ is the QLoRA block, with $\mathbf{A} \in \mathbb{R}^{d\times r}$ the down projection, $\mathbf{B}^T \in \mathbb{R}^{r\times d}$ the up projection, $\alpha$ a scaling factor, and $r$ the rank. 

$\mathbf{A}$ is initialised with the Kaiming uniform distribution, and $\mathbf{B}$ with all-zero initialisation.
$\Delta\Theta$ is optimised with a standard conditional language modelling objective with CE loss computed over $\Theta_0 + \Delta\Theta$ (i.e.\ the frozen model with QLoRA blocks plugged in). For a given input $x$, the output $h$ resulting from adapting the frozen parameters $\mathbf{W_0}$ with a QLoRA block is: 

\vspace{-.3cm}
\[h = \mathbf{W_0}\:x + \frac{\alpha}{r} \mathbf{\Delta W}x\]
\vspace{-.5cm}

\noindent In our implementation of plug-and-play QLoRA, multiple blocks can be attached in each location and composed in one of three ways as detailed next.

\subsection{Composition techniques}\label{sec:composition-techniques}

\begin{figure}
    \centering
    \includegraphics[width=1\linewidth]{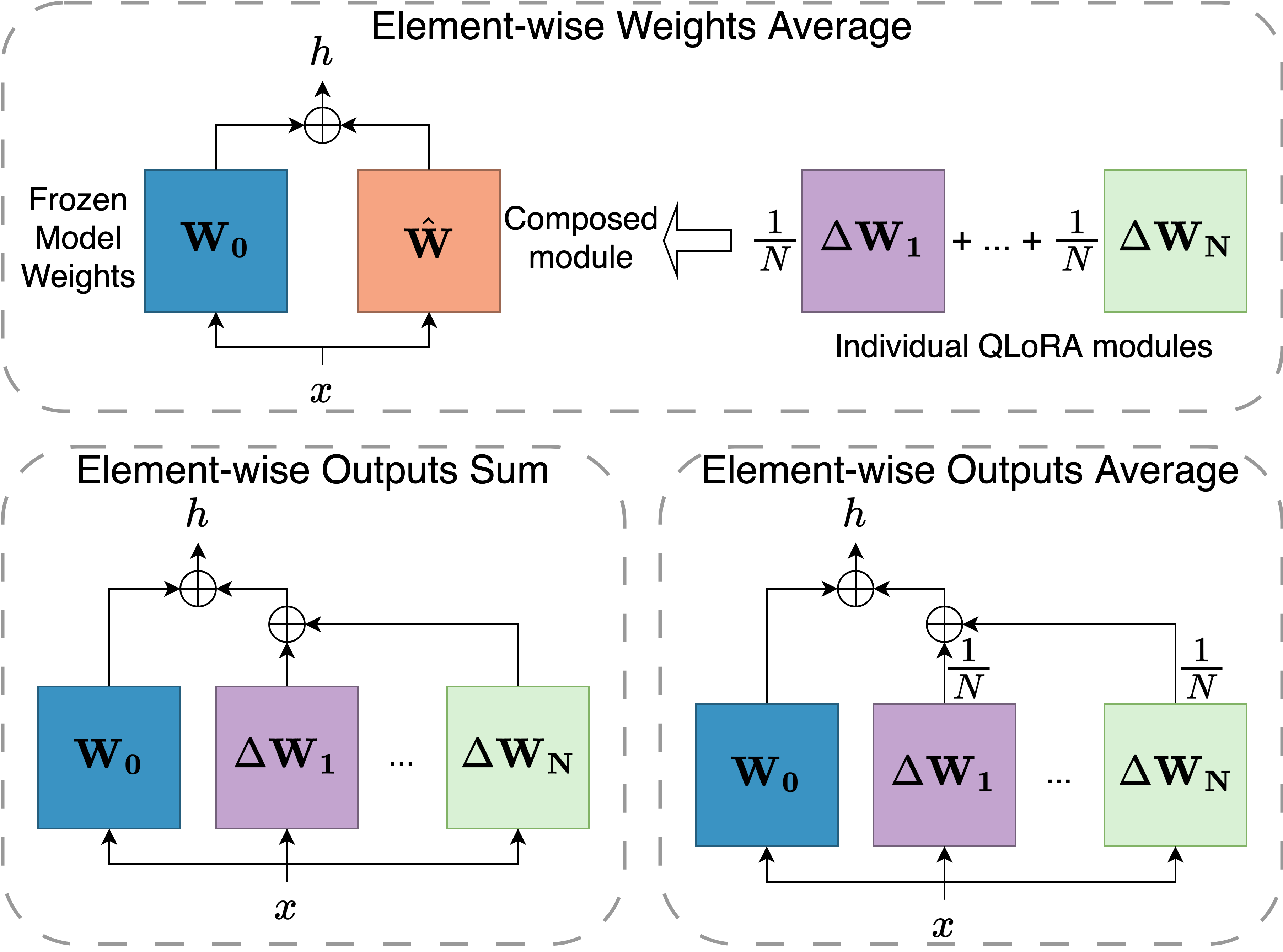}
    \caption{Diagram of our three QLoRA module composition techniques: (Top) Element-wise Weights Average; (Bottom-left) Element-wise Outputs Summing; (Bottom-right) Element-wise Outputs Averaging. }
    \label{fig:mod_combination}
\end{figure}

Figure~\ref{fig:mod_combination} shows the three composition techniques we tested in diagrammatic form. The first is \textbf{weights averaging}, 
where $N$ PEFT modules are composed by computing the element-wise average of their weights (see top of Figure~\ref{fig:mod_combination}): 

\vspace{-.3cm}
\[\mathbf{\hat{W}} = \frac{1}{N}\sum_{i=1}^{N}\Delta \mathbf{W_i}\]

\noindent where $\Delta \mathbf{W_i}$ are the weights of the $i$-th QLoRA module, and $\mathbf{\hat{W}}$ is the single module resulting from the composition. The output from the latter is summed to the output from the frozen parameters:

\vspace{-.3cm}
\[h = \mathbf{W_0}\:x + \frac{\alpha}{r} \mathbf{\hat{W}}x\]

\vspace{-.2cm}

\noindent The second composition technique is \textbf{output summing}, where the outputs of the $N$ modules to be composed are first scaled by their respective scaling factors ($\frac{\alpha}{r}=\frac{16}{64}$ in all tests, see Section~\ref{sec:train}), and the element-wise sum of outputs (Figure~\ref{fig:mod_combination}, bottom left) is then computed and added to $\mathbf{W_0}\:x$:
\vspace{-.3cm}
\[h = \mathbf{W_0}\:x + \sum_{i=1}^{N}\frac{\alpha_i}{r_i} (\Delta \mathbf{W_i}\:x)\]

Output summing scales the total adapter contribution by the number of modules $N$. In our experiments we compose 2--5 modules; if tasks require substantially more modules to be composed, there may be effects on performance that we cannot predict from the experiments reported here. 

Our third technique, \textbf{output averaging},  computes the element-wise average of module outputs (Figure~\ref{fig:mod_combination}, bottom right), before adding it to $\mathbf{W_0}x$:
\vspace{-.3cm}
\[h = \mathbf{W_0}\:x + \frac{1}{N}\sum_{i=1}^{N}\frac{\alpha_i}{r_i} (\Delta \mathbf{W_i}\:x)\]

Note that weight averaging and output averaging are not mathematically equivalent as further explained in Appendix~\ref{app:weight_vs_output}. 

The theoretical advantage of output composition over weight composition lies in preserving learned module structure. Weight averaging merges the weight matrices themselves, losing the individual low-rank structures encoding task-specific information. Output summing computes each module's output before combining them, preserving task-specific representational directions that can be additively integrated in activation space, as shown by representation engineering findings (Section~\ref{sec:rel_research}). 

Our approach requires no additional training, but inference cost scales linearly with the number of modules $N$: output summing and averaging require $N$ adapter forward passes for $N$ modules. Weight averaging maintains the same proportional inference cost but has consistently weaker performance (Section~\ref{sec:res}), presenting a trade-off between computational efficiency and control effectiveness.

\section{Study Overview}
\label{sec:study_overview}

We aim to assess if the above composition techniques can achieve performant results on each of the individual module tasks, as well as on composite tasks requiring their combined functionality. We train QLoRA modules in conjunction with the frozen host model on individual tasks, compose them with one of the above three techniques, then test them on both individual and composite tasks. 

For each pretrained raw model $\Theta$, individual datasets $\mathbf{D} = \{D_1, ... D_m\}$, and composition techniques $\mathbf{C} = \{C_1, ... C_p\}$, we report results in terms of the schema shown in Table~\ref{tab:study-struc} which gives the results tables (Tables~\ref{tab:sent_res}, \ref{tab:topic_res}, and \ref{tab:multi_res}) their structure and number of rows (\#rows).

\begin{table}[h!]
    \centering\small
\setlength\tabcolsep{2pt} % default value: 6pt
\renewcommand{\arraystretch}{1.20}
\begin{tabular}{|l|c|}
\hline
QLoRA-tuned model (composition)& $\#$rows\\
    \hline
    Raw model $\Theta$ & 1 \\
    + $\Delta\Theta_{D_i}$ trained on $D_i$ & $m$ \\
    + $\Delta\Theta_{\mathbf{D}}$ trained on combined $D_1 \cup ... D_m$ & 1 \\
    \hdashline
    + $C_1$ applied to subsets of  $\Delta\Theta_{D_1}$, ...$\Delta\Theta_{D_m}$ & $\sum_{k=2}^{m}$ $ m\choose k$\\
    \dottedline
    + $C_2$ applied to subsets of  $\Delta\Theta_{D_1}$, ...$\Delta\Theta_{D_m}$ & $\sum_{k=2}^{m}$ $ m\choose k$\\
    \dottedline
    + $...$ & {}\\
    \dottedline
    + $C_p$ applied to subsets of  $\Delta\Theta_{D_1}$, ...$\Delta\Theta_{D_m}$ & $\sum_{k=2}^{m}$ $ m\choose k$\\
    \hline
\end{tabular}
    \caption{Study structure in terms of raw model and QLoRA (composed) modules. $D_i$ = individual datasets; $C_i$ = composition techniques. For other details, see text.}
    \label{tab:study-struc}
\end{table}

This study structure provides multiple points of comparison in terms of the performance of (i) the raw model on each dataset; (ii) the raw model + QLoRA module finetuned on each of the individual datasets, tested (a) on the data set it was finetuned on, and (b) on other related datasets; (iii) the raw model + QLoRA module finetuned on all individual datasets combined, tested on each separate data set; and (iv) the raw model + different types of composition applied to separately finetuned QLoRA modules, tested on the individual data sets.

\section{Models and Datasets}\label{sec:model_data}

\textbf{Models.} We use three raw pretrained LLMs as host models: LLaMa 3 8B~\citep{llama3modelcard}, LLaMa 3.1 8B, and Mistral 7B~\citep{jiang2023mistral}. For full details of models, see Appendix~\ref{app:models}.

\textbf{Datasets.} We use two sets of  widely used single-task datasets, three for sentiment analysis which we use for sentiment-controlled text generation; and two  for topic detection which we use for topic-controlled text generation. The \textbf{\textit{sentiment}} datasets are Yelp Reviews~\citep{chelba2013one} (short review texts labelled pos/neg/neu), Stanford Sentiment Treebank (SST-2)~\citep{socher-etal-2013-recursive} (sentences extracted from reviews labelled with the sentence-level label pos/neg), and IMDb Movie Reviews~\citep{maas-EtAl:2011:ACL-HLT2011} (longer review texts labelled pos/neg). The \textit{\textbf{topic}} datasets are AG News \citep{NIPS2015_250cf8b5} (short news summaries labelled  Sport, Business, Science/Tech, World); and DBpedia \citep{NIPS2015_250cf8b5} (very short entity descriptions with 14 entity class labels which we map to the AG News topics (as per Appendix~\ref{app:data}).

We use two short-text datasets for out-of-domain testing: PPLM Prompts \cite{dathathri2019plug}, a collection of 35 2--4 word prompts, randomly chosen from sentence starters recommended for academic writing;\footnote{\url{http://www2.eit.ac.nz/library/ls_guides_sentencestarters.html}} and the single-sentence image captions from the STS benchmark test set \citep{cer-etal-2017-semeval} from which we derive prompts in two ways: (i) using the whole sentence as a prompt (\textit{STS} in tables), and (ii) using just the first $n$ words as prompts, $n=0..5$ (\textit{STS proc} in tables). Neither of these datasets has attribute labels, so we generate and test one text for each control attribute label (combination) for each prompt in the dataset. 

\begin{table*}[h!]
    \centering
    \scriptsize
    \setlength\tabcolsep{2pt} % default value: 6pt
    \renewcommand{\arraystretch}{1.10}
    \begin{tabular}{|l|ccc|c|c|cccc|cccc|}
    \hline
         \multirow{3}{*}{\textbf{CTG Technique}} & \multicolumn{3}{c|}{\multirow{2}{*}{\textbf{Distinct-n}$\uparrow$}} & \multirow{3}{*}{\textbf{SLOR}$\uparrow$} & \multicolumn{9}{c|}{\textbf{Control Effectiveness}$\uparrow$} \\ \cline{6-14}
         & & & & & \multirow{2}{*}{\textbf{Avg All}} & \multirow{2}{*}{\textbf{Avg}} & \multirow{2}{*}{\textbf{Yelp}} & \multirow{2}{*}{\textbf{IMDB}} & \multirow{2}{*}{\textbf{SST-2}} & \multicolumn{4}{c|}{\textbf{Out-Of-Domain}} \\ \cline{2-4} \cline{11-14}
         & \textbf{dist-1} & \textbf{dist-2} & \textbf{dist-3} &  &  &  &  &  &  & \textbf{Avg} & \textbf{PPLM S} & \textbf{STS S} & \textbf{STS proc S} \\
\hline \hline
Llama 3 8B & 0.03 & 0.09 & 0.12 & 9.44 & 59.93 & 63.70 & 68.00 & 57.56 & 65.56 & 57.16 & 64.60 & 52.83 & 54.06 \\
+ QLoRA Yelp & 0.09 & 0.24 & 0.34 & 9.87 & \cellcolor{gray!25}85.11 & 91.44 & \cellcolor{gray!25}\underline{92.78} & 90.67 & 90.89 & \cellcolor{gray!25}79.49 & 88.41 & \textbf{67.78} & \cellcolor{gray!25}82.28 \\
+ QLoRA IMDB & 0.04 & 0.13 & 0.19 & \textbf{10.86} & 83.07 & 92.07 & 92.11 & \underline{89.67} & \cellcolor{gray!25}94.44 & 73.61 & 80.95 & 62.00 & 77.89 \\
+ QLoRA SST-2 & \textbf{0.34} & \textbf{0.64} & \textbf{0.71} & 8.24 & 77.80 & 85.96 & 84.11 & 82.44 & \underline{91.33} & 70.73 & 82.86 & 57.39 & 71.94 \\
+ QLoRA Combined Sentiment dataset& 0.11 & 0.27 & 0.35 & 10.07 & 82.99 & 91.07 & \underline{92.22} & \cellcolor{gray!25}\underline{92.00} & \underline{89.00} & 75.88 & 87.46 & 63.50 & 76.67 \\
\hdashline
+ QLoRA Output Summing(IMDB, SST-2) & 0.15 & 0.36 & 0.46 & 9.43 & 82.12 & 90.85 & 89.78 & 89.56 & 93.22 & 74.85 & \cellcolor{gray!25}88.89 & 61.72 & 73.94 \\
+ QLoRA Output Summing(IMDB, Yelp) & 0.07 & 0.21 & 0.31 & 10.41 & 83.98 & 91.22 & 90.22 & 89.33 & 94.11 & 77.49 & 87.46 & 64.44 & 80.56 \\
+ QLoRA Output Summing(Yelp, SST-2) & 0.21 & 0.46 & 0.56 & 8.93 & 81.98 & 90.74 & 92.44 & 90.22 & 89.56 & 74.96 & \textbf{90.00} & 58.33 & 76.56 \\
+ QLoRA Output Summing(IMDB, Yelp, SST-2) & 0.13 & 0.34 & 0.45 & 9.83 & \textbf{85.66} & \textbf{92.48} & \textbf{94.44} & \textbf{92.56} & 90.44 & \textbf{79.54} & \cellcolor{gray!25}88.89 & \cellcolor{gray!25}67.00 & \textbf{82.72} \\
\dottedline
+ QLoRA Output Averaging(IMDB, SST-2) & 0.12 & 0.27 & 0.35 & 9.49 & 79.29 & 88.30 & 86.33 & 87.44 & 91.11 & 71.24 & 83.65 & 58.44 & 71.61 \\
+ QLoRA Output Averaging(IMDB, Yelp) & 0.05 & 0.16 & 0.23 & 10.30 & 83.70 & \cellcolor{gray!25}92.26 & 91.11 & 89.67 & \textbf{96.00} & 75.67 & 86.19 & 63.17 & 77.67 \\
+ QLoRA Output Averaging(Yelp, SST-2) & 0.13 & 0.31 & 0.38 & 9.16 & 81.96 & 89.74 & 89.00 & 88.56 & 91.67 & 74.69 & 84.29 & 62.61 & 77.17 \\
+ QLoRA Output Averaging(IMDB, Yelp, SST-2) & 0.07 & 0.19 & 0.27 & 9.77 & 81.04 & 89.81 & 88.78 & 89.00 & 91.67 & 73.03 & 84.60 & 60.50 & 74.00 \\
\dottedline
+ QLoRA Averaged Weights(IMDB, SST-2) & 0.31 & \cellcolor{gray!25}0.61 & \cellcolor{gray!25}0.69 & 8.37 & 77.50 & 85.93 & 85.33 & 81.89 & 90.56 & 69.16 & 77.94 & 57.50 & 72.06 \\
+ QLoRA Averaged Weights(IMDB, Yelp) & 0.04 & 0.12 & 0.19 & \cellcolor{gray!25}10.83 & 81.96 & 90.74 & 91.89 & 87.22 & 93.11 & 73.65 & 84.13 & 60.67 & 76.17 \\
+ QLoRA Averaged Weights(Yelp, SST-2) & \cellcolor{gray!25}0.33 & \textbf{0.64} & \textbf{0.71} & 8.28 & 77.94 & 85.93 & 85.33 & 81.89 & 90.56 & 71.44 & 84.76 & 57.50 & 72.06 \\
+ QLoRA Averaged Weights(IMDB, Yelp, SST-2) & 0.11 & 0.28 & 0.36 & 9.54 & 80.29 & 90.04 & 88.89 & 88.78 & 92.44 & 71.17 & 83.17 & 58.72 & 71.61 \\
\hline \hline
Llama 3.1 8B & 0.04 & 0.10 & 0.13 & 9.66 & 58.32 & 61.52 & 66.22 & 54.22 & 64.11 & 56.41 & 64.29 & 51.56 & 53.39 \\
+ QLoRA Yelp & 0.09 & 0.26 & 0.36 & 9.94 & \cellcolor{gray!25}86.68 & 91.52 & \underline{91.00} & 89.00 & 94.56 & \textbf{82.93} & \cellcolor{gray!25}91.75 & \textbf{71.28} & \textbf{85.78} \\
+ QLoRA IMDB & 0.04 & 0.14 & 0.21 & \textbf{10.81} & 83.46 & 92.26 & 89.67 & \cellcolor{gray!25}\underline{91.44} & \textbf{95.67} & 75.04 & 85.24 & 63.33 & 76.56 \\
+ QLoRA SST-2 & \textbf{0.36} & \textbf{0.69} & \textbf{0.76} & 8.24 & 78.29 & 85.48 & 82.33 & 82.89 & \underline{91.22} & 72.49 & 84.76 & 58.17 & 74.56 \\
+ QLoRA Combined Sentiment  dataset & 0.12 & 0.27 & 0.35 & 10.00 & 84.75 & 92.67 & \cellcolor{gray!25}\underline{94.89} & \cellcolor{gray!25}\underline{91.44} & \underline{91.67} & 77.86 & 89.52 & 62.28 & 81.78 \\
\hdashline
+ QLoRA Output Summing(IMDB, SST-2) & 0.09 & 0.23 & 0.31 & 9.91 & 84.98 & \cellcolor{gray!25}92.93 & 93.44 & 90.89 & 94.44 & 78.68 & \textbf{92.54} & 62.78 & 80.72 \\
+ QLoRA Output Summing(IMDB, Yelp) & 0.07 & 0.21 & 0.31 & 10.33 & 85.50 & 91.52 & 91.22 & 90.00 & 93.33 & 80.26 & 89.05 & 68.06 & 83.67 \\
+ QLoRA Output Summing(Yelp, SST-2) & 0.16 & 0.38 & 0.47 & 9.16 & 85.38 & 92.48 & 92.78 & \cellcolor{gray!25}91.44 & 93.22 & 79.07 & 89.05 & \cellcolor{gray!25}68.94 & 79.22 \\
+ QLoRA Output Summing(IMDB, Yelp, SST-2) & 0.10 & 0.27 & 0.37 & 9.91 & \textbf{87.35} & \textbf{94.70} & \textbf{96.33} & \textbf{92.67} & \cellcolor{gray!25}95.11 & \cellcolor{gray!25}80.61 & 90.16 & 66.89 & \cellcolor{gray!25}84.78 \\
\dottedline
+ QLoRA Output Averaging(IMDB, SST-2) & 0.08 & 0.21 & 0.28 & 9.81 & 81.14 & 89.74 & 89.22 & 87.22 & 92.78 & 72.95 & 83.02 & 61.50 & 74.33 \\
+ QLoRA Output Averaging(IMDB, Yelp) & 0.07 & 0.22 & 0.31 & 10.24 & 83.45 & 92.30 & 93.00 & 90.22 & 93.67 & 75.16 & 86.03 & 63.22 & 76.22 \\
+ QLoRA Output Averaging(Yelp, SST-2) & 0.16 & 0.36 & 0.44 & 9.05 & 80.54 & 89.00 & 89.67 & 86.89 & 90.44 & 73.06 & 85.08 & 60.94 & 73.17 \\
+ QLoRA Output Averaging(IMDB, Yelp, SST-2) & 0.07 & 0.18 & 0.26 & 10.01 & 79.79 & 89.15 & 88.56 & 87.11 & 91.78 & 71.69 & 85.56 & 58.39 & 71.11 \\
\dottedline
+ QLoRA Averaged Weights(IMDB, SST-2) & 0.33 & 0.64 & 0.71 & 8.38 & 76.87 & 84.07 & 81.00 & 81.89 & 89.33 & 71.21 & 83.97 & 56.67 & 73.00 \\
+ QLoRA Averaged Weights(IMDB, Yelp) & 0.04 & 0.14 & 0.21 & \cellcolor{gray!25}10.78 & 82.24 & 91.19 & 90.67 & 88.89 & 94.00 & 73.57 & 83.49 & 61.89 & 75.33 \\
+ QLoRA Averaged Weights(Yelp, SST-2) & \cellcolor{gray!25}0.35 & \cellcolor{gray!25}0.67 & \cellcolor{gray!25}0.74 & 8.27 & 76.83 & 84.07 & 81.00 & 81.89 & 89.33 & 71.00 & 83.33 & 56.67 & 73.00 \\
+ QLoRA Averaged Weights(IMDB, Yelp, SST-2) & 0.08 & 0.20 & 0.27 & 9.87 & 81.48 & 89.67 & 91.67 & 86.67 & 90.67 & 74.19 & 85.56 & 60.72 & 76.28 \\
\hline \hline
Mistral 7B & 0.04 & 0.09 & 0.12 & 7.45 & 57.54 & 60.00 & 59.33 & 55.00 & 65.67 & 56.23 & 62.86 & 52.50 & 53.33 \\
+ QLoRA Yelp & 0.03 & 0.07 & 0.10 & 10.44 & \textbf{84.33} & 92.30 & \underline{\textbf{94.78}} & \textbf{90.56} & 91.56 & \textbf{76.42} & 84.60 & \textbf{61.33} & \textbf{83.33} \\
+ QLoRA IMDB & 0.02 & 0.04 & 0.06 & \textbf{11.25} & 80.24 & 90.63 & 89.67 & \underline{88.56} & 93.67 & 70.38 & 82.70 & 58.78 & 69.67 \\
+ QLoRA SST-2 & \textbf{0.27} & \textbf{0.51} & \textbf{0.58} & 7.65 & 78.17 & 85.67 & 83.11 & 81.67 & \underline{92.22} & 71.87 & 83.65 & 55.94 & 76.00 \\
+ QLoRA Combined Sentiment  dataset & 0.03 & 0.06 & 0.09 & 10.83 & 78.28 & 86.30 & \underline{87.11} & \underline{85.67} & \underline{86.11} & 71.32 & 83.17 & 58.22 & 72.56 \\
\hdashline
+ QLoRA Output Summing(IMDB, SST-2) & 0.22 & 0.41 & 0.47 & 7.80 & 79.28 & 87.11 & 85.56 & 85.44 & 90.33 & 73.12 & \cellcolor{gray!25}86.98 & 57.11 & 75.28 \\
+ QLoRA Output Summing(IMDB, Yelp) & 0.03 & 0.07 & 0.10 & 10.51 & 82.53 & 90.44 & 93.33 & 86.33 & 91.67 & \cellcolor{gray!25}75.59 & \cellcolor{gray!25}86.98 & 58.28 & \cellcolor{gray!25}81.50 \\
+ QLoRA Output Summing(Yelp, SST-2) & 0.18 & 0.35 & 0.40 & 8.22 & 80.17 & 88.26 & 89.00 & 82.33 & 93.44 & 74.13 & \textbf{89.68} & 59.22 & 73.50 \\
+ QLoRA Output Summing(IMDB, Yelp, SST-2) & 0.08 & 0.16 & 0.19 & 9.13 & \cellcolor{gray!25}83.06 & \textbf{92.81} & \cellcolor{gray!25}93.78 & 89.44 & \textbf{95.22} & 74.09 & 86.67 & 58.89 & 76.72 \\
\dottedline
+ QLoRA Output Averaging(IMDB, SST-2) & 0.11 & 0.21 & 0.25 & 8.49 & 79.09 & 88.89 & 88.56 & 85.33 & 92.78 & 69.93 & 82.06 & 58.39 & 69.33 \\
+ QLoRA Output Averaging(IMDB, Yelp) & 0.02 & 0.06 & 0.09 & 10.80 & 82.23 & \cellcolor{gray!25}92.63 & 93.56 & \cellcolor{gray!25}90.11 & 94.22 & 72.35 & 84.60 & 56.78 & 75.67 \\
+ QLoRA Output Averaging(Yelp, SST-2) & 0.05 & 0.10 & 0.13 & 8.84 & 80.80 & 90.74 & 92.44 & 86.33 & 93.44 & 71.96 & 85.87 & \cellcolor{gray!25}60.67 & 69.33 \\
+ QLoRA Output Averaging(IMDB, Yelp, SST-2) & 0.02 & 0.05 & 0.07 & 10.54 & 81.31 & 92.37 & 91.67 & \textbf{90.56} & \cellcolor{gray!25}94.89 & 70.99 & 84.76 & 57.67 & 70.56 \\
\dottedline
+ QLoRA Averaged Weights(IMDB, SST-2) & \cellcolor{gray!25}0.25 & 0.49 & 0.55 & 7.69 & 78.22 & 85.85 & 82.89 & 82.22 & 92.44 & 71.98 & 84.60 & 56.28 & 75.06 \\
+ QLoRA Averaged Weights(IMDB, Yelp) & 0.02 & 0.04 & 0.07 & \cellcolor{gray!25}11.19 & 80.71 & 91.33 & 90.22 & 89.89 & 93.89 & 70.72 & 83.65 & 59.56 & 68.94 \\
+ QLoRA Averaged Weights(Yelp, SST-2) & \textbf{0.27} & \cellcolor{gray!25}0.50 & \cellcolor{gray!25}0.57 & 7.66 & 78.35 & 85.85 & 82.89 & 82.22 & 92.44 & 72.61 & 86.51 & 56.28 & 75.06 \\
+ QLoRA Averaged Weights(IMDB, Yelp, SST-2) & 0.10 & 0.20 & 0.24 & 8.64 & 79.19 & 88.26 & 88.11 & 84.11 & 92.56 & 71.67 & 86.35 & 58.00 & 70.67 \\
\hline
    \end{tabular}
    \caption{\textbf{Sentiment Control} Diversity, Fluency, Control Effectiveness for the model + QLoRA module combinations explained in Table~\ref{tab:study-struc}. Here, e.g.\ Output Summing(data1, data2) refers to the output summation module composition technique. All values are averages over 3 runs. Standard deviations are reported in Appendix Tables~\ref{tab:sent_res_std_dist} and \ref{tab:sent_res_std}. Bold (shaded) =  (second) highest score in column/section; underline = train/test on same dataset.} 
    \label{tab:sent_res}
\end{table*}

\section{Experimental Set-up}
\label{sec:method}

\vspace{-.05cm}
\subsection{Task construal}
\vspace{-.05cm}

We use the above in-domain datasets for finetuning QLoRA modules on attribute-controlled prompted text generation. Prompts are leading text fragments tagged with the target attribute value(s), and the output is the text generated by the model in response, up to the first end-of-output tag, or a specified maximum length, whichever comes first. 

Leading fragments are the first $n$ words of a given text, where $n$ ranges from 0 to 5 (2--4 for PPLM). Input-output pairs are then constructed by splitting the tagged text after $n$ words, e.g.\ for sentiment-controlled generation control:

\vspace{0.2cm}
\begin{small}\begin{tabular}{p{0.9\linewidth}}\texttt{[SENTIMENT] Positive [$\backslash$SENTIMENT] [ANS] The sushi is} 

\vspace{0.1cm}

\texttt{great! Not expensive \& good quality. My favorite rolles are the Vegas, Dragon, and the baked scallops :) [$\backslash$ANS]}\\
\end{tabular}\end{small}
\vspace{0.05cm}

\vspace{-.05cm}

\vspace{-.05cm}
\subsection{QLoRA module training}\label{sec:train}
\vspace{-.05cm}

As a baseline, we finetune QLoRA modules on composite datasets constructed by filtering, balancing, and stratified sampling of multiple datasets from Section~\ref{sec:model_data} to ensure size and label distribution consistency (see Appendix~\ref{app:data} for details).

A small grid search is performed for each model with the goal to balance affordable compute with model performance. Following the vanilla QLoRA hyperparameters \cite{dettmers2023qlora}, we set rank to 64, alpha to 16 and dropout to 0.1. QLoRA matrices are initialised using the Kaiming uniform distribution for A, using $a=\sqrt{5}$, and all-zero initialisation for B (see also Section~\ref{sec:plug-play-qlora}). We train all our QLoRA modules in half-precision on an NVIDIA A100 with 80GB for 3 epochs. We save a checkpoint at each epoch, and select the one with the best Control Effectiveness (see Section~\ref{sec:eval}) on the validation set.
Each training/testing run is performed three times with different seeds.

We train one QLoRA module each for the five individual (Section~\ref{sec:model_data}) and two composite datasets (Section~\ref{sec:sampling} in Appendix).

\vspace{-.05cm}
\subsection{Testing}
\vspace{-.05cm}

We test the three composition techniques from Section~\ref{sec:composition-techniques} as per the study structure from Table~\ref{tab:study-struc}, for two properties: (i) generalisation over related tasks, which here means all \textbf{sentiment control} tasks, or all \textbf{topic control} tasks; and (ii) functional composability, which here means the ability to control both sentiment and topic via module composition. The latter is a \textbf{multiple-attribute control} task, and here we test two distinct composition strategies: (i) composing the two single-attribute modules trained on the composite Combined Sentiment/Topic datasets; and (ii) composing the five single-attribute modules trained on the separate individual datasets. 

\begin{table*}[h]
    \centering
    \scriptsize
    \setlength\tabcolsep{2pt} % default value: 6pt
    \renewcommand{\arraystretch}{1.10}
    \begin{tabular}{|l|ccc|c|c|ccc|cccc|}
    \hline
         \multirow{3}{*}{\textbf{CTG Technique}} & \multicolumn{3}{c|}{\multirow{2}{*}{\textbf{Distinct-n}$\uparrow$}} & \multirow{3}{*}{\textbf{SLOR}$\uparrow$} & \multicolumn{8}{c|}{\textbf{Control Effectiveness}$\uparrow$} \\ \cline{6-13}
         & & & & & \multirow{2}{*}{\textbf{Avg All}} & \multirow{2}{*}{\textbf{Avg}} & \multirow{2}{*}{\textbf{AG News}} & \multirow{2}{*}{\textbf{DBPedia}} & \multicolumn{4}{c|}{\textbf{Out-Of-Domain}} \\ \cline{2-4} \cline{10-13}
         & \textbf{dist-1} & \textbf{dist-2} & \textbf{dist-3} &  &  &  &  &  & \textbf{Avg} & \textbf{PPLM T} & \textbf{STS T} & \textbf{STS proc T} \\
\hline
Llama 3 8B & 0.07 & 0.17 & 0.22 & 8.45 & 45.47 & 58.52 & 64.61 & 52.42 & 37.53 & 48.97 & 27.97 & 35.64 \\
+ QLoRA AG News & 0.25 & 0.53 & 0.62 & \cellcolor{gray!25}9.39 & \textbf{68.53} & \textbf{85.13} & \underline{\textbf{90.72}} & \textbf{79.54} & \textbf{58.02} & \textbf{71.11} & \textbf{42.03} & \cellcolor{gray!25}60.92 \\
+ QLoRA DBPedia & 0.32 & 0.60 & 0.68 & 8.96 & 52.61 & 69.94 & 71.67 & \underline{68.21} & 41.66 & 55.40 & 28.58 & 41.00 \\
+ QLoRA Combined Topic dataset & 0.30 & 0.60 & 0.70 & 8.92 & 62.76 & 74.42 & \underline{81.61} & \underline{67.24} & \cellcolor{gray!25}56.27 & \cellcolor{gray!25}68.73 & \cellcolor{gray!25}37.50 & \textbf{62.58} \\
\hdashline
+ QLoRA Output Summing(AG News, DBPedia) & \textbf{0.35} & \textbf{0.70} & \textbf{0.80} & 9.33 & \cellcolor{gray!25}63.61 & \cellcolor{gray!25}82.57 & \cellcolor{gray!25}88.78 & \cellcolor{gray!25}76.35 & 52.59 & \textbf{71.11} & 32.81 & 53.86 \\
+ QLoRA Output Averaging(AG News, DBPedia) & 0.27 & 0.55 & 0.64 & \textbf{9.57} & 60.97 & 78.33 & 83.78 & 72.88 & 50.70 & 66.98 & 33.81 & 51.31 \\
+ QLoRA Averaged Weights(AG News, DBPedia) & \cellcolor{gray!25}0.33 & \cellcolor{gray!25}0.62 & \cellcolor{gray!25}0.71 & 8.98 & 53.90 & 70.61 & 72.11 & 69.12 & 45.50 & 66.59 & 28.06 & 41.86 \\
\hline \hline
Llama 3.1 8B & 0.05 & 0.12 & 0.17 & \textbf{9.45} & 45.86 & 58.61 & 66.11 & 51.11 & 37.47 & 46.35 & 29.89 & 36.17 \\
+ QLoRA AG News & 0.26 & 0.55 & 0.65 & \cellcolor{gray!25}9.40 & \textbf{67.85} & \textbf{85.52} & \underline{\textbf{91.33}} & \textbf{79.72} & \textbf{57.21} & \textbf{73.10} & \cellcolor{gray!25}36.28 & \textbf{62.25} \\
+ QLoRA DBPedia & 0.31 & 0.59 & 0.68 & 8.74 & 52.63 & 69.86 & 70.94 & \underline{68.77} & 41.64 & 54.92 & 28.11 & 41.89 \\
+ QLoRA Combined Topic dataset & 0.31 & \cellcolor{gray!25}0.62 & \cellcolor{gray!25}0.73 & 9.29 & \cellcolor{gray!25}65.08 & 80.65 & \cellcolor{gray!25}\underline{89.06} & \underline{72.25} & \cellcolor{gray!25}55.01 & 66.51 & \textbf{39.97} & \cellcolor{gray!25}58.56 \\
\hdashline
+ QLoRA Output Summing(AG News, DBPedia) & \textbf{0.35} & \textbf{0.68} & \textbf{0.78} & 9.12 & 63.41 & \cellcolor{gray!25}82.91 & 88.39 & \cellcolor{gray!25}77.44 & 51.64 & \cellcolor{gray!25}69.05 & 29.39 & 56.47 \\
+ QLoRA Output Averaging(AG News, DBPedia) & 0.30 & 0.59 & 0.68 & 8.98 & 58.85 & 76.42 & 83.44 & 69.40 & 47.74 & 61.67 & 29.81 & 51.75 \\
+ QLoRA Averaged Weights(AG News, DBPedia) & \cellcolor{gray!25}0.32 & 0.61 & 0.71 & 8.74 & 52.59 & 69.52 & 70.94 & 68.09 & 43.06 & 60.71 & 27.61 & 40.86 \\
\hline \hline
Mistral 7B & 0.07 & 0.13 & 0.17 & 7.98 & 44.14 & 55.96 & 62.00 & 49.91 & 37.48 & 49.76 & 29.92 & 32.75 \\
+ QLoRA AG News & 0.18 & 0.37 & 0.44 & \cellcolor{gray!25}9.48 & \textbf{69.05} & \textbf{88.52} & \underline{\textbf{93.17}} & \textbf{83.87} & \textbf{57.14} & \textbf{73.97} & \textbf{39.69} & \textbf{57.75} \\
+ QLoRA DBPedia & \textbf{0.26} & \cellcolor{gray!25}0.48 & \cellcolor{gray!25}0.56 & 8.81 & 52.92 & 67.05 & 67.50 & \underline{66.61} & 43.76 & 54.13 & 32.89 & 44.28 \\
+ QLoRA Combined Topic dataset& \cellcolor{gray!25}0.21 & 0.43 & 0.52 & 9.09 & 64.64 & 81.74 & \cellcolor{gray!25}\underline{89.00} & \underline{74.47} & \cellcolor{gray!25}53.87 & 67.54 & \cellcolor{gray!25}39.22 & \cellcolor{gray!25}54.86 \\
\hdashline
+ QLoRA Output Summing(AG News, DBPedia) & \textbf{0.26} & \textbf{0.50} & \textbf{0.59} & 9.26 & \cellcolor{gray!25}65.46 & \cellcolor{gray!25}85.27 & 88.89 & \cellcolor{gray!25}81.65 & 53.19 & \cellcolor{gray!25}69.76 & 36.36 & 53.44 \\
+ QLoRA Output Averaging(AG News, DBPedia) & 0.18 & 0.36 & 0.42 & \textbf{9.60} & 58.01 & 76.01 & 82.00 & 70.03 & 46.64 & 60.95 & 33.89 & 45.08 \\
+ QLoRA Averaged Weights(AG News, DBPedia) & \textbf{0.26} & \cellcolor{gray!25}0.48 & \cellcolor{gray!25}0.56 & 8.87 & 53.58 & 66.53 & 66.06 & 67.01 & 46.72 & 61.75 & 33.47 & 44.94 \\
\hline
    \end{tabular}
    \caption{Diversity, Fluency, Control Effectiveness for \textbf{Topic Control}, training on \textit{single} and \textit{combined} datasets, and composition of modules trained on single datasets, e.g.\ Output Summing(data1, data2). All values are averages over 3 runs. Standard deviations are reported in Appendix Tables~\ref{tab:topic_res_std_dist} and \ref{tab:topic_res_std_ce}. Bold (shaded) =  (second) highest score in column and section; underline = train and test set from same dataset.}
    \label{tab:topic_res}
\end{table*}

\vspace{-.05cm}
\subsection{Evaluation metrics}\label{sec:eval}
\vspace{-.05cm}

We measure \textit{\textbf{Diversity}} with Distinct-n~\citep{li2015diversity} which is the proportion of unique n-grams in generated texts. The aggregate Distinct-n score is the mean of item-level scores.

We assess \textit{\textbf{Fluency}} with average Syntactic Log-Odds Ratio (SLOR)~\citep{kann-etal-2018-sentence} obtained with GPT-2XL and BLOOM 1B7, in preference over perplexity which may not effectively capture fluency for low-frequency items. SLOR calculates the log-probability of a sentence normalised by unigram log-probability and length. 

To assess \textit{\textbf{Control Effectiveness}} (CE), we measure the mean percentage of texts identified by a set of classifiers as possessing the controlled attribute value. I.e.\ we first compute the percentage of cases for each classifier where the detected attribute value matches the input control value; the final CE score is then the average of the three classifiers' individual percentages.
We use DistilBERT and T5 fine-tuned on SST-2, and DeBERTa fine-tuned on Yelp, for sentiment control; and DistilBERT, BERT, and DeBERTa fine-tuned on AG-News, for topic control (Appendix~\ref{app:eval_metr}). In multiple-attribute control, instead of one target attribute value having to be matched, both have to be matched. We use majority voting across classifiers to obtain sentiment and topic labels before calculating CE as above.

\section{Results}
\label{sec:res}

\begin{table*}[h]
    \centering
    \scriptsize
    \setlength\tabcolsep{1pt} % default value: 6pt
    \renewcommand{\arraystretch}{1.10}
    \begin{tabular}{|l|ccc|c|c|ccc|ccc|ccc|}
\hline
\multirow{4}{*}{\textbf{CTG Technique}} & \multicolumn{3}{c|}{\multirow{2}{*}{\textbf{Distinct-n}$\uparrow$}} & \multirow{4}{*}{\textbf{SLOR}$\uparrow$} & \multicolumn{10}{c|}{\textbf{Control Effectiveness} $\uparrow$} \\ \cline{6-15}
&  &  &  & &  \multicolumn{10}{c|}{\textbf{Out-Of-Domain}} \\ \cline{6-15}
& \multirow{2}{*}{\textbf{dist1}} & \multirow{2}{*}{\textbf{dist2}} & \multirow{2}{*}{\textbf{dist3}} & & \multicolumn{4}{c|}{\textbf{Multiple}} & \multicolumn{3}{c|}{\textbf{Sentiment}} & \multicolumn{3}{c|}{\textbf{Topic}} \\
\cline{6-15}
&  &  &  & & \textbf{Avg} & \textbf{PPLM M} & \textbf{STS M} & \textbf{STS p M} &\textbf{PPLM S} & \textbf{STS S} & \textbf{STS p S} & \textbf{PPLM T} & \textbf{STS T} & \textbf{STS p T} \\
\hline
Llama 3 8B & 0.04 & 0.11 & 0.16 & 9.77 & 19.22 & 29.29 & 14.54 & 20.38 & 64.60 & 52.83 & 54.06 & 48.97 & 27.97 & 35.64 \\
+ QLoRA Combined Sentiment  dataset & 0.03 & 0.09 & 0.14 & \textbf{10.81} & \textbf{25.66} & \cellcolor{gray!25}37.14 & 16.50 & \textbf{30.79} & \cellcolor{gray!25}87.46 & \textbf{63.50} & \textbf{76.67} & 51.67 & 27.69 & 38.36 \\
+ QLoRA Combined Topic dataset & \textbf{0.24} & \textbf{0.50} & \textbf{0.59} & 8.52 & 22.64 & 36.55 & 14.96 & 25.46 & 55.71 & 50.78 & 54.83 & \textbf{68.73} & \textbf{37.50} & \textbf{62.58} \\
\hdashline
+ QLoRA Output Summing(S, T) & 0.09 & 0.23 & 0.31 & \cellcolor{gray!25}9.81 & 23.85 & 35.60 & \textbf{17.75} & 25.83 & \textbf{87.62} & \cellcolor{gray!25}59.56 & \cellcolor{gray!25}72.72 & 62.46 & 29.42 & 50.25 \\
+ QLoRA Output Summing(Ind mod) & 0.18 & 0.42 & 0.51 & 9.11 & \cellcolor{gray!25}24.08 & 33.69 & \cellcolor{gray!25}16.79 & \cellcolor{gray!25}28.00 & 79.68 & 59.22 & 69.39 & 43.97 & 25.22 & 37.75 \\
\dottedline
+ QLoRA Output Averaging(S, T) & 0.10 & 0.24 & 0.31 & 9.26 & 22.32 & 36.55 & 14.50 & 25.17 & 76.35 & 54.56 & 63.83 & 57.86 & 29.39 & 42.11 \\
+ QLoRA Output Averaging(Ind mod) & 0.08 & 0.20 & 0.26 & 9.55 & 23.30 & \textbf{37.50} & 15.46 & 26.17 & 78.41 & 57.17 & 67.61 & 57.94 & 26.58 & 43.22 \\
\dottedline
+ QLoRA Averaged Weights(S, T) & \cellcolor{gray!25}0.21 & \cellcolor{gray!25}0.46 & \cellcolor{gray!25}0.55 & 8.56 & 22.85 & 36.55 & 15.62 & 25.29 & 74.92 & 50.78 & 54.83 & \cellcolor{gray!25}67.70 & \cellcolor{gray!25}36.83 & \cellcolor{gray!25}59.86 \\
+ QLoRA Averaged Weights(Ind mod) & 0.09 & 0.23 & 0.30 & 9.45 & 23.58 & 35.71 & 15.79 & 27.12 & 73.65 & 54.50 & 63.17 & 64.84 & 32.11 & 46.33 \\
\hline \hline
Llama 3.1 8B & 0.05 & 0.12 & 0.18 & \cellcolor{gray!25}9.84 & 19.95 & 28.57 & 15.00 & 21.88 & 64.29 & 51.56 & 53.39 & 46.35 & 29.89 & 36.17 \\
+ QLoRA Combined Sentiment  dataset & 0.03 & 0.11 & 0.17 & \textbf{10.73} & \cellcolor{gray!25}26.44 & 37.50 & \cellcolor{gray!25}17.54 & \textbf{31.46} & \textbf{89.52} & \cellcolor{gray!25}62.28 & \textbf{81.78} & 49.13 & 30.72 & 36.86 \\
+ QLoRA Combined Topic dataset & \textbf{0.28} & \textbf{0.58} & \textbf{0.68} & 8.58 & 23.95 & 35.12 & \cellcolor{gray!25}17.54 & 26.46 & 52.06 & 51.39 & 51.83 & \cellcolor{gray!25}66.51 & \textbf{39.97} & \cellcolor{gray!25}58.56 \\
\hdashline
+ QLoRA Output Summing(S, T) & 0.10 & 0.27 & 0.36 & 9.80 & \textbf{26.76} & 34.40 & \textbf{19.58} & \cellcolor{gray!25}31.25 & 75.56 & \textbf{62.89} & 71.83 & 58.17 & 34.97 & 49.75 \\
+ QLoRA Output Summing(Ind mod) & 0.13 & 0.29 & 0.35 & 8.63 & 20.78 & 30.24 & 16.29 & 21.96 & \cellcolor{gray!25}84.60 & 60.94 & \cellcolor{gray!25}72.94 & 35.00 & 24.42 & 28.92 \\
\dottedline
+ QLoRA Output Averaging(S, T) & 0.11 & 0.29 & 0.37 & 9.67 & 25.89 & \textbf{40.83} & 16.50 & 30.04 & 73.81 & 59.83 & 63.72 & 59.44 & 29.61 & 43.08 \\
+ QLoRA Output Averaging(Ind mod) & 0.08 & 0.22 & 0.29 & 9.36 & 23.17 & 37.38 & 15.21 & 26.17 & 69.84 & 54.50 & 64.89 & 45.87 & 27.03 & 39.42 \\
\dottedline
+ QLoRA Averaged Weights(S, T) & \cellcolor{gray!25}0.24 & \cellcolor{gray!25}0.52 & \cellcolor{gray!25}0.62 & 8.74 & 24.43 & 37.98 & 17.29 & 26.83 & 73.49 & 51.39 & 51.83 & \textbf{70.08} & \cellcolor{gray!25}39.61 & \textbf{61.11} \\
+ QLoRA Averaged Weights(Ind mod) & 0.10 & 0.25 & 0.33 & 9.14 & 23.26 & \cellcolor{gray!25}40.36 & 15.46 & 25.08 & 72.22 & 53.61 & 64.06 & 50.40 & 28.28 & 39.81 \\
\hline \hline
Mistral 7B & 0.03 & 0.07 & 0.10 & 9.77 & 20.43 & 26.07 & 16.12 & 22.75 & 62.86 & 52.50 & 53.33 & 49.76 & 29.92 & 32.75 \\
+ QLoRA Combined Sentiment  dataset & 0.01 & 0.03 & 0.05 & \textbf{11.60} & 21.47 & 30.60 & 14.75 & 25.00 & \cellcolor{gray!25}83.17 & \textbf{58.22} & \cellcolor{gray!25}72.56 & 46.83 & 27.44 & 38.33 \\
+ QLoRA Combined Topic dataset & \textbf{0.12} & \textbf{0.27} & \textbf{0.36} & 8.92 & \cellcolor{gray!25}26.06 & 35.12 & \cellcolor{gray!25}20.33 & \cellcolor{gray!25}28.62 & 50.16 & 51.50 & 52.33 & \cellcolor{gray!25}67.54 & 39.22 & 54.86 \\
\hdashline
+ QLoRA Output Summing(S, T) & 0.09 & 0.22 & 0.30 & 9.50 & 25.53 & 34.05 & 19.58 & 28.50 & 59.84 & 51.33 & 57.56 & 63.33 & \cellcolor{gray!25}39.75 & \textbf{60.88} \\
+ QLoRA Output Summing(Ind mod) & 0.09 & 0.20 & 0.24 & 8.94 & 22.77 & \cellcolor{gray!25}36.19 & 16.25 & 24.58 & \textbf{87.78} & \cellcolor{gray!25}57.33 & \textbf{75.83} & 57.30 & 31.44 & 42.14 \\
\dottedline
+ QLoRA Output Averaging(S, T) & 0.08 & 0.18 & 0.23 & 9.63 & 22.73 & 34.88 & 15.54 & 25.67 & 63.17 & 52.83 & 57.17 & 60.87 & 29.92 & 43.97 \\
+ QLoRA Output Averaging(Ind mod) & 0.02 & 0.05 & 0.08 & \cellcolor{gray!25}10.75 & 23.01 & 35.95 & 14.08 & 27.42 & 76.19 & 54.78 & 66.89 & 48.17 & 27.81 & 37.61 \\
\dottedline
+ QLoRA Averaged Weights(S, T) & \cellcolor{gray!25}0.10 & \cellcolor{gray!25}0.24 & \cellcolor{gray!25}0.32 & 9.04 & \textbf{26.24} & 27.14 & \textbf{21.25} & \textbf{30.92} & 59.05 & 51.50 & 52.33 & \textbf{70.00} & \textbf{39.92} & \cellcolor{gray!25}55.78 \\
+ QLoRA Averaged Weights(Ind mod) & 0.03 & 0.07 & 0.10 & 10.18 & 21.70 & \textbf{36.31} & 14.42 & 23.88 & 77.46 & 54.28 & 62.44 & 52.86 & 28.00 & 38.78 \\
\hline
\end{tabular}
\caption{Diversity, Fluency, Control Effectiveness for \textbf{Multi-attribute Control} alongside single-attribute control results for comparison. All values are averages over 3 runs. Standard deviations are reported in Appendix Tables~\ref{tab:multi_res_std_dist} and \ref{tab:multi_res_std_ce}. S=the Combined Sentiment dataset, T=the Combined Topic dataset, M=Multiple, p = processed, Ind mod=composition is on all 5 individually trained modules. Bold (shaded) =  (second) highest score in column and section; underline = train and test set from same dataset.} 
\label{tab:multi_res}
\end{table*}

\vspace{-.05cm}
\subsection{Sentiment control}
\vspace{-.05cm}

Table~\ref{tab:sent_res} presents results for sentiment control in terms of the row structure introduced in Table~\ref{tab:study-struc}. Across the columns, we have results for Diversity measured by Distinct-n, Fluency by SLOR, and Control Effectiveness by classifier average (see Section~\ref{sec:eval}), on in-domain and out-of-domain datasets (Section~\ref{sec:model_data}), as well as averaged over each. 

Regarding \textit{Diversity}, strikingly, training on SST-2 always achieves the most diverse outputs by a very substantial margin, across all three models, although Weights Averaging closely matches this if the SST-2 trained module is in the composition. For \textit{Fluency}, it is training on IMDb that achieves the highest scores, albeit by smaller margins.

For \textit{Control Effectiveness}, while the three (unchanged) raw models perform on average on a par (first row in each section), Mistral responds slightly worse to  QLoRA-tuning across all settings. We observe a clear trend where Output Summing  consistently achieves the highest scores for module compositions, except for Mistral, where results present a slightly more mixed picture. Despite this, the Summing technique achieves best or second-best results in out-of-domain testing across all models.

Regarding individual train/test set combinations, strikingly it is rarely the case that the model trained on a single dataset achieves the best result on it. For the Llama models, when training on the combined data, performance is improved or maintained compared to training on the same data set in all cases but one (SST-2). In contrast, for Mistral, performance always drops considerably.

For the last three columns (out-of-domain testing), the Yelp-trained model versions always achieve the best result, in most cases by considerable margins, with the overall best results achieved by Llama 3.1 8B + QLoRA Yelp.

These results present a clear overall trend where our new output composition approaches (Summing and Averaging) consistently outperform the Weights Average method across all models and evaluation settings, indicating that combining the outputs of separately fine-tuned PEFT modules is a more effective strategy for achieving control effectiveness than averaging their weights.

\subsection{Topic control}

Table~\ref{tab:topic_res} presents results for topic control using the same structure and notation as in Table~\ref{tab:sent_res}. \textit{Diversity} scores here are generally higher than for sentiment. Summing the two modules performs best for all settings. \textit{Fluency} scores are very similar across the different settings with no notable trends emerging. 

In \textit{Control Effectiveness}, for both  in-domain and out-of-domain datasets, the single AG News trained module (second row in each  section) performs best in all but two cases. The summed modules have the second highest scores in 11 cases, and the single module trained on the Combined dataset has the second highest scores also in 11 cases. 

Among the composition techniques, we confirm the consistent trend where Output Summing outperforms the other methods.

Note that training on AG News has a twofold advantage over DBPedia in the current context: (i) the classifiers in the CE metric were all trained on AG News, and (ii) the mapping from topics in DBPedia to AG News labels is imperfect resulting in a noisier text-label relationship. To assess the impact we conducted an additional evaluation using an instruction-tuned LLM as a classifier which confirmed the findings in Table~\ref{tab:topic_res}, with both Pearson’s and Spearman's correlations around or above 0.9 (see Appendix~\ref{app:llm_classifier} for details). This suggests no adverse effect from classifier bias.

\subsection{Multiple-attribute control}

The single-attribute results so far presented shed light on the ability of module composition to generalise across individual datasets. The results for multiple-attribute control over both sentiment and topic presented in this section test the functional composability of the modules.

Table~\ref{tab:multi_res} presents the multiple-attribute control results. We can see that in terms of \textit{Diversity}, modules trained on (just) the Combined Topic dataset achieve the highest results for all models. In terms of \textit{Fluency}, all models perform best with the module trained on the Combined Sentiment dataset.

For \textit{Control Effectiveness} overall, finetuning  consistently performs above raw model baseline for the Llama models, by big margins. For Mistral, finetuned models in some cases perform worse.

For Multiple-control, results present a mixed picture. One perspective is provided by comparing modules trained on the Combined datasets with those composing modules trained on individual datasets. Here, results are dataset and model dependent: for Llama, Output Summing is best for STS M, Output Averaging is best for PPLM M, and Combined dataset training is best for STS p M; for Mistral generally, Weights Averaging is best.

The finetuned models in this section were all task-specifically created for multiple-attribute control, so the above are the main results in this section. However, we also wanted to see to what extent the multi-attribute control models retain their single-attribute control performance, shown in the last six columns in Table~\ref{tab:multi_res}. A clear trend is that best results are worse than in single-attribute control (Tables~\ref{tab:sent_res} and \ref{tab:topic_res}). Which composition methods work best also follows clear trends: for retaining sentiment control performance, it is the module trained on the Combined Topic dataset combined with Weights Averaging. For retaining topic control performance, it is the module trained on the Combined Sentiment dataset combined with Output Summing. 

For retention of topic control performance,  Weights Averaging  yields  best or second-best results across all models. However, with Mistral, our new Summing approach achieves comparable or higher performance than Weights Averaging, suggesting that output-based composition is also competitive in this scenario under certain conditions.

\section{Discussion}
\label{sec:disc}

State-of-the-art controlled-generation methods like Prior CTG~\cite{gu-etal-2023-controllable}, PPLM~\cite{dathathri2019plug}, and GeDi~\cite{krause-etal-2021-gedi-generative} report control effectiveness in the range of 80–97\% and 74–97\% for sentiment and topic control, compared to our ranges of 86--96\% and 70--93\%, respectively. For multi-attribute control, average single-attribute performance tends to be reported, with results typically in the range 71--92\%. This is clearly not comparable to our results where we use three classifiers and require all predicted labels to be correct. More comparably, for MacLaSa, \citet{ding-etal-2023-maclasa} compute CE as the percentage of times both predicted labels are correct, reporting considerably lower effectiveness, ranging from 18–59\%, better reflecting the difficulty of controlling multiple attributes. 

Across all experiments, Output Summing consistently outperforms other composition techniques. Our results indicate that summing the output of multiple PEFT modules effectively preserves the knowledge captured in individual modules, particularly when composing modules trained on the same control attribute. The success of summing (and to a lesser extent, averaging) module outputs suggests that trained modules project inputs into the same latent space effectively. 

When combining modules trained on different tasks or control attributes, we observe that composed modules successfully preserve single-attribute knowledge, particularly for sentiment. However, topic control is less effective, suggesting that the signal from topic modules may be less strong, leading to weaker control over diverse attributes. A possible improvement could be weighted module composition, where specific attributes are reinforced to ensure stronger control. Despite this, our findings confirm that even when composing modules that have been fine-tuned for different single-attribute tasks, the model retains its single-attribute control capabilities. 

Three-module output summing outperforms task-specialized single modules in 7 of 9 scenarios, demonstrating that composition not only preserves but enhances single-task performance through cross-dataset generalization (see Appendix~\ref{sec:app_mod_comp} for detailed analysis).

\section{Conclusion}
\label{sec:concl}

We have presented a first-time examination of the output composability of multiple QLoRA modules within the same host model, in a fully plug-and-play setting. We set out to assess two tests of composability: (i) generalisation over related tasks, and (ii) functional composability on composite tasks. Re \textit{i}, our new output summing composition method consistently provided the best performance, not only generalising well over multiple tasks, but even improving performance on the individual tasks, averaging a 2\% \textit{improvement} over single-task trained modules on the corresponding single task test set. Re \textit{ii}, module composition overall provided an advantage, but could not always outperform training a single module on data set combinations. Our results provide a first demonstration of the astonishing composability of QLoRA-finetuned modules.

\section*{Limitations}
\label{sec:limit}
In this work, we focus on exploring our approach only on three pre-trained raw models with similar number of parameters, same number of layers and same architecture type. While this choice allows us to isolate and test the effects of our approach, it restricts our understanding of its general applicability. Testing our approach on models with different number of parameters, architecture type, and number of layers would give a more comprehensive picture of its robustness and versatility.

Our evaluation is limited to two control attributes, sentiment and topic, across related datasets within the same domain. While this focused scope allows us to systematically isolate composition mechanics, it limits the generality of our conclusions. Whether the composability patterns we observe, particularly the effectiveness of output summing, extend to (i) more complex generative behaviours such as reasoning or creative writing, (ii) fundamentally different task types (e.g., combining classification, question-answering, and generation modules), or (iii) unrelated domains remains unexplored. These represent important directions for future work to establish the broader applicability of PEFT module composition.

Our experiments compose 2-5 modules trained on related tasks within the same domain. Real-world applications may require orchestrating dozens or hundreds of modules across truly disparate tasks (e.g., combining modules for translation, summarization, code generation, and question-answering). Such large-scale scenarios introduce several challenges we do not address: (1) computational cost scaling linearly with active modules (O(N)) could become prohibitive; (2) output summing scales adapter contributions linearly with N, potentially causing output magnitudes to deviate significantly from training distributions; (3) potential interference effects between unrelated task modules; and (4) determining which modules to activate for complex queries. Addressing large-scale heterogeneous composition would likely require additional mechanisms such as: dynamic module selection or mixture-of-experts-style routing to activate only relevant subsets; normalization or learned weighting schemes to modulate individual contributions and control output magnitude; adaptive scaling factors; or meta-learning to predict module compatibility and optimal composition strategies.

Additionally, our stratified sampling approach ensures all modules are trained on equal-sized datasets, which enables fair comparison of composition techniques but does not explore how output summing performs with varying training data sizes, an important consideration for real-world scenarios where modules may be trained on datasets of vastly different scales.

Furthermore, we limit this work to investigating only one PEFT technique, which leaves open the question of whether our approach would perform the same with other PEFT techniques. Expanding the scope of our work to include different PEFT techniques could help evaluate the generality of the approach. Furthermore, incorporating the composability of modules generated by different PEFT techniques would shed light on the extent to which our method is compatible with mixed-module designs.

Finally, our evaluation setup relies on automatic metrics to assess the model's performance. A dedicated human evaluation would provide a more nuanced understanding of the generated texts.

\section*{Ethical Considerations}
\label{sec:ethic}
We test our approach on two well-known control attributes, namely sentiment and topic. However, our approach could potentially be used to force the model to include and generate offensive, inappropriate, or biased content. At present, our approach does not include any built-in mechanisms to prevent or mitigate such misuse, highlighting an important area for future work in ensuring ethical and responsible application.

Additionally, our approach is based on the use of pre-trained LLMs, which are known to inherit biases and limitations from the data they were trained on. The generated outputs might include offensive, incorrect, or harmful content.

\section*{Acknowledgments}

Michela Lorandi’s work was conducted with the financial support of the Science Foundation Ireland Centre for Research Training in Digitally-Enhanced Reality (d-real) under Grant No. 18/CRT/6224. Both authors benefit from being members of the ADAPT SFI Research Centre at Dublin City University, funded by the Science Foundation Ireland under Grant Agreement No. 13/RC/2106\_P2. For the purpose of Open Access, the author has applied a CC BY public copyright licence to any Author Accepted Manuscript version arising from this submission.

\bibliography{custom}

@inproceedings{
zhao2025merging,
title={Merging Lo{RA}s like Playing {LEGO}: Pushing the Modularity of Lo{RA} to Extremes Through Rank-Wise Clustering},
author={Ziyu Zhao and Tao Shen and Didi Zhu and Zexi Li and Jing Su and Xuwu Wang and Fei Wu},
booktitle={The Thirteenth International Conference on Learning Representations},
year={2025},
url={https://openreview.net/forum?id=j6fsbpAllN}
}

@article{turner2023activation,
  title={Activation Addition: Steering Language Models Without Optimization},
  author={Turner, Alexander Matt and Thiergart, Lisa and Udell, David and Leech, Gavin and Mini, Ulisse and MacDiarmid, Monte},
  journal={CoRR},
  year={2023}
}

@article{zou2023representation,
  title={Representation Engineering: A Top-Down Approach to AI Transparency},
  author={Zou, Andy and Phan, Long and Chen, Sarah and Campbell, James and Guo, Phillip and Ren, Richard and Pan, Alexander and Yin, Xuwang and Mazeika, Mantas and Dombrowski, Ann-Kathrin and others},
  journal={CoRR},
  year={2023}
}

@article{li2023inference,
  title={Inference-time intervention: Eliciting truthful answers from a language model},
  author={Li, Kenneth and Patel, Oam and Vi{\'e}gas, Fernanda and Pfister, Hanspeter and Wattenberg, Martin},
  journal={Advances in Neural Information Processing Systems},
  volume={36},
  pages={41451--41530},
  year={2023}
}

@inproceedings{ding-etal-2023-maclasa,
    title = "{M}ac{L}a{S}a: Multi-Aspect Controllable Text Generation via Efficient Sampling from Compact Latent Space",
    author = "Ding, Hanxing  and
      Pang, Liang  and
      Wei, Zihao  and
      Shen, Huawei  and
      Cheng, Xueqi  and
      Chua, Tat-Seng",
    editor = "Bouamor, Houda  and
      Pino, Juan  and
      Bali, Kalika",
    booktitle = "Findings of the Association for Computational Linguistics: EMNLP 2023",
    month = dec,
    year = "2023",
    address = "Singapore",
    publisher = "Association for Computational Linguistics",
    url = "https://aclanthology.org/2023.findings-emnlp.292/",
    doi = "10.18653/v1/2023.findings-emnlp.292",
    pages = "4424--4436"
}

@inproceedings{krause-etal-2021-gedi-generative,
    title = "{G}e{D}i: Generative Discriminator Guided Sequence Generation",
    author = "Krause, Ben  and
      Gotmare, Akhilesh Deepak  and
      McCann, Bryan  and
      Keskar, Nitish Shirish  and
      Joty, Shafiq  and
      Socher, Richard  and
      Rajani, Nazneen Fatema",
    editor = "Moens, Marie-Francine  and
      Huang, Xuanjing  and
      Specia, Lucia  and
      Yih, Scott Wen-tau",
    booktitle = "Findings of the Association for Computational Linguistics: EMNLP 2021",
    month = nov,
    year = "2021",
    address = "Punta Cana, Dominican Republic",
    publisher = "Association for Computational Linguistics",
    url = "https://aclanthology.org/2021.findings-emnlp.424/",
    doi = "10.18653/v1/2021.findings-emnlp.424",
    pages = "4929--4952"
}

@article{loshchilov2017decoupled,
  title={Decoupled weight decay regularization},
  author={Loshchilov, I},
  journal={arXiv preprint arXiv:1711.05101},
  year={2017}
}

@inproceedings{dou-etal-2024-loramoe,
    title = "{L}o{RAM}o{E}: Alleviating World Knowledge Forgetting in Large Language Models via {M}o{E}-Style Plugin",
    author = "Dou, Shihan  and
      Zhou, Enyu  and
      Liu, Yan  and
      Gao, Songyang  and
      Shen, Wei  and
      Xiong, Limao  and
      Zhou, Yuhao  and
      Wang, Xiao  and
      Xi, Zhiheng  and
      Fan, Xiaoran  and
      Pu, Shiliang  and
      Zhu, Jiang  and
      Zheng, Rui  and
      Gui, Tao  and
      Zhang, Qi  and
      Huang, Xuanjing",
    editor = "Ku, Lun-Wei  and
      Martins, Andre  and
      Srikumar, Vivek",
    booktitle = "Proceedings of the 62nd Annual Meeting of the Association for Computational Linguistics (Volume 1: Long Papers)",
    month = aug,
    year = "2024",
    address = "Bangkok, Thailand",
    publisher = "Association for Computational Linguistics",
    url = "https://aclanthology.org/2024.acl-long.106/",
    doi = "10.18653/v1/2024.acl-long.106",
    pages = "1932--1945"
}

@article{huang2023lorahub,
  title={Lorahub: Efficient cross-task generalization via dynamic lora composition},
  author={Huang, Chengsong and Liu, Qian and Lin, Bill Yuchen and Pang, Tianyu and Du, Chao and Lin, Min},
  journal={arXiv preprint arXiv:2307.13269},
  year={2023}
}

@inproceedings{poth2023adapters,
  title={Adapters: A Unified Library for Parameter-Efficient and Modular Transfer Learning},
  author={Poth, Clifton and Sterz, Hannah and Paul, Indraneil and Purkayastha, Sukannya and Engl{\"a}nder, Leon and Imhof, Timo and Vuli{\'c}, Ivan and Ruder, Sebastian and Gurevych, Iryna and Pfeiffer, Jonas},
  booktitle={Proceedings of the 2023 Conference on Empirical Methods in Natural Language Processing: System Demonstrations},
  pages={149--160},
  year={2023}
}

@inproceedings{feng-etal-2024-mixture,
    title = "Mixture-of-{L}o{RA}s: An Efficient Multitask Tuning Method for Large Language Models",
    author = "Feng, Wenfeng  and
      Hao, Chuzhan  and
      Zhang, Yuewei  and
      Han, Yu  and
      Wang, Hao",
    editor = "Calzolari, Nicoletta  and
      Kan, Min-Yen  and
      Hoste, Veronique  and
      Lenci, Alessandro  and
      Sakti, Sakriani  and
      Xue, Nianwen",
    booktitle = "Proceedings of the 2024 Joint International Conference on Computational Linguistics, Language Resources and Evaluation (LREC-COLING 2024)",
    month = may,
    year = "2024",
    address = "Torino, Italia",
    publisher = "ELRA and ICCL",
    url = "https://aclanthology.org/2024.lrec-main.994/",
    pages = "11371--11380"
}

@inproceedings{asadi2024combining,
  title={Combining Pre-trained LoRA Modules Improves Few-shot Adaptation of Foundation Models to New Tasks},
  author={Asadi, Nader and Beitollahi, Mahdi and Khalil, Yasser H and Li, Yinchuan and Zhang, Guojun and Chen, Xi},
  booktitle={ICML 2024 Workshop on Foundation Models in the Wild},
    year={2024}
}

@article{hu2021lora,
  title={Lora: Low-rank adaptation of large language models},
  author={Hu, Edward J and Shen, Yelong and Wallis, Phillip and Allen-Zhu, Zeyuan and Li, Yuanzhi and Wang, Shean and Wang, Lu and Chen, Weizhu},
  journal={arXiv preprint arXiv:2106.09685},
  year={2021}
}

@inproceedings{whitehouse-etal-2024-low,
    title = "Low-Rank Adaptation for Multilingual Summarization: An Empirical Study",
    author = "Whitehouse, Chenxi  and
      Huot, Fantine  and
      Bastings, Jasmijn  and
      Dehghani, Mostafa  and
      Lin, Chu-Cheng  and
      Lapata, Mirella",
    editor = "Duh, Kevin  and
      Gomez, Helena  and
      Bethard, Steven",
    booktitle = "Findings of the Association for Computational Linguistics: NAACL 2024",
    month = jun,
    year = "2024",
    address = "Mexico City, Mexico",
    publisher = "Association for Computational Linguistics",
    url = "https://aclanthology.org/2024.findings-naacl.77/",
    doi = "10.18653/v1/2024.findings-naacl.77",
    pages = "1202--1228"
}

@article{zhao2024lora,
  title={LoRA Land: 310 Fine-tuned LLMs that Rival GPT-4, A Technical Report},
  author={Zhao, Justin and Wang, Timothy and Abid, Wael and Angus, Geoffrey and Garg, Arnav and Kinnison, Jeffery and Sherstinsky, Alex and Molino, Piero and Addair, Travis and Rishi, Devvret},
  journal={arXiv preprint arXiv:2405.00732},
  year={2024}
}

@article{jiang2023mistral,
  title={Mistral 7B},
  author={Jiang, Albert Q and Sablayrolles, Alexandre and Mensch, Arthur and Bamford, Chris and Chaplot, Devendra Singh and Casas, Diego de las and Bressand, Florian and Lengyel, Gianna and Lample, Guillaume and Saulnier, Lucile and others},
  journal={arXiv preprint arXiv:2310.06825},
  year={2023}
}

@article{liu2024gpt,
  title={GPT understands, too},
  author={Liu, Xiao and Zheng, Yanan and Du, Zhengxiao and Ding, Ming and Qian, Yujie and Yang, Zhilin and Tang, Jie},
  journal={AI Open},
  volume={5},
  pages={208--215},
  year={2024},
  publisher={Elsevier}
}

@inproceedings{NIPS2015_250cf8b5,
 author = {Zhang, Xiang and Zhao, Junbo and LeCun, Yann},
 booktitle = {Advances in Neural Information Processing Systems},
 editor = {C. Cortes and N. Lawrence and D. Lee and M. Sugiyama and R. Garnett},
 pages = {},
 publisher = {Curran Associates, Inc.},
 title = {Character-level Convolutional Networks for Text Classification},
 url = {https://proceedings.neurips.cc/paper_files/paper/2015/file/250cf8b51c773f3f8dc8b4be867a9a02-Paper.pdf},
 volume = {28},
 year = {2015}
}

@InProceedings{maas-EtAl:2011:ACL-HLT2011,
  author    = {Maas, Andrew L.  and  Daly, Raymond E.  and  Pham, Peter T.  and  Huang, Dan  and  Ng, Andrew Y.  and  Potts, Christopher},
  title     = {Learning Word Vectors for Sentiment Analysis},
  booktitle = {Proceedings of the 49th Annual Meeting of the Association for Computational Linguistics: Human Language Technologies},
  month     = {June},
  year      = {2011},
  address   = {Portland, Oregon, USA},
  publisher = {Association for Computational Linguistics},
  pages     = {142--150},
  url       = {http://www.aclweb.org/anthology/P11-1015}
}

@article{llama3modelcard,
title={Llama 3 Model Card},
author={AI@Meta},
year={2024},
url = {https://github.com/meta-llama/llama3/blob/main/MODEL_CARD.md}
}

@article{dettmers2023qlora,
  title={QLoRA: Efficient Finetuning of Quantized LLMs},
  author={Dettmers, Tim and Pagnoni, Artidoro and Holtzman, Ari and Zettlemoyer, Luke},
  journal={arXiv preprint arXiv:2305.14314},
  year={2023}
}

@inproceedings{cer-etal-2017-semeval,
    title = "{S}em{E}val-2017 Task 1: Semantic Textual Similarity Multilingual and Crosslingual Focused Evaluation",
    author = "Cer, Daniel  and
      Diab, Mona  and
      Agirre, Eneko  and
      Lopez-Gazpio, I{\~n}igo  and
      Specia, Lucia",
    editor = "Bethard, Steven  and
      Carpuat, Marine  and
      Apidianaki, Marianna  and
      Mohammad, Saif M.  and
      Cer, Daniel  and
      Jurgens, David",
    booktitle = "Proceedings of the 11th International Workshop on Semantic Evaluation ({S}em{E}val-2017)",
    month = aug,
    year = "2017",
    address = "Vancouver, Canada",
    publisher = "Association for Computational Linguistics",
    url = "https://aclanthology.org/S17-2001",
    doi = "10.18653/v1/S17-2001",
    pages = "1--14",
}

@article{dathathri2019plug,
  title={Plug and play language models: A simple approach to controlled text generation},
  author={Dathathri, Sumanth and Madotto, Andrea and Lan, Janice and Hung, Jane and Frank, Eric and Molino, Piero and Yosinski, Jason and Liu, Rosanne},
  journal={arXiv preprint arXiv:1912.02164},
  year={2019}
}

@inproceedings{sabry-belz-2024-assessing,
    title = "Assessing the Portability of Parameter Matrices Trained by Parameter-Efficient Finetuning Methods",
    author = "Sabry, Mohammed  and
      Belz, Anya",
    editor = "Graham, Yvette  and
      Purver, Matthew",
    booktitle = "Findings of the Association for Computational Linguistics: EACL 2024",
    month = mar,
    year = "2024",
    address = "St. Julian{'}s, Malta",
    publisher = "Association for Computational Linguistics",
    url = "https://aclanthology.org/2024.findings-eacl.106/",
    pages = "1548--1556"
}

@InProceedings{pmlr-v97-houlsby19a,
  title = 	 {Parameter-Efficient Transfer Learning for {NLP}},
  author =       {Houlsby, Neil and Giurgiu, Andrei and Jastrzebski, Stanislaw and Morrone, Bruna and De Laroussilhe, Quentin and Gesmundo, Andrea and Attariyan, Mona and Gelly, Sylvain},
  booktitle = 	 {Proceedings of the 36th International Conference on Machine Learning},
  pages = 	 {2790--2799},
  year = 	 {2019},
  editor = 	 {Chaudhuri, Kamalika and Salakhutdinov, Ruslan},
  volume = 	 {97},
  series = 	 {Proceedings of Machine Learning Research},
  month = 	 {09--15 Jun},
  publisher =    {PMLR},
  url = 	 {https://proceedings.mlr.press/v97/houlsby19a.html}
}

@inproceedings{li-liang-2021-prefix,
    title = "Prefix-Tuning: Optimizing Continuous Prompts for Generation",
    author = "Li, Xiang Lisa  and
      Liang, Percy",
    booktitle = "Proceedings of the 59th Annual Meeting of the Association for Computational Linguistics and the 11th International Joint Conference on Natural Language Processing (Volume 1: Long Papers)",
    month = aug,
    year = "2021",
    address = "Online",
    publisher = "Association for Computational Linguistics",
    url = "https://aclanthology.org/2021.acl-long.353",
    doi = "10.18653/v1/2021.acl-long.353",
    pages = "4582--4597",
}

@article{li2015diversity,
  title={A diversity-promoting objective function for neural conversation models},
  author={Li, Jiwei and Galley, Michel and Brockett, Chris and Gao, Jianfeng and Dolan, Bill},
  journal={arXiv preprint arXiv:1510.03055},
  year={2015}
}

@inproceedings{kann-etal-2018-sentence,
    title = "Sentence-Level Fluency Evaluation: References Help, But Can Be Spared!",
    author = "Kann, Katharina  and
      Rothe, Sascha  and
      Filippova, Katja",
    editor = "Korhonen, Anna  and
      Titov, Ivan",
    booktitle = "Proceedings of the 22nd Conference on Computational Natural Language Learning",
    month = oct,
    year = "2018",
    address = "Brussels, Belgium",
    publisher = "Association for Computational Linguistics",
    url = "https://aclanthology.org/K18-1031",
    doi = "10.18653/v1/K18-1031",
    pages = "313--323",
}

@inproceedings{gu-etal-2023-controllable,
    title = "Controllable Text Generation via Probability Density Estimation in the Latent Space",
    author = "Gu, Yuxuan  and
      Feng, Xiaocheng  and
      Ma, Sicheng  and
      Zhang, Lingyuan  and
      Gong, Heng  and
      Zhong, Weihong  and
      Qin, Bing",
    editor = "Rogers, Anna  and
      Boyd-Graber, Jordan  and
      Okazaki, Naoaki",
    booktitle = "Proceedings of the 61st Annual Meeting of the Association for Computational Linguistics (Volume 1: Long Papers)",
    month = jul,
    year = "2023",
    address = "Toronto, Canada",
    publisher = "Association for Computational Linguistics",
    url = "https://aclanthology.org/2023.acl-long.704",
    doi = "10.18653/v1/2023.acl-long.704",
    pages = "12590--12616",
}

@inproceedings{socher-etal-2013-recursive,
    title = "Recursive Deep Models for Semantic Compositionality Over a Sentiment Treebank",
    author = "Socher, Richard  and
      Perelygin, Alex  and
      Wu, Jean  and
      Chuang, Jason  and
      Manning, Christopher D.  and
      Ng, Andrew  and
      Potts, Christopher",
    booktitle = "Proceedings of the 2013 Conference on Empirical Methods in Natural Language Processing",
    month = oct,
    year = "2013",
    address = "Seattle, Washington, USA",
    publisher = "Association for Computational Linguistics",
    url = "https://www.aclweb.org/anthology/D13-1170",
    pages = "1631--1642",
}

@inproceedings{chelba2013one,
  title={One billion word benchmark for measuring progress in statistical language modeling},
  author={Chelba, Ciprian and Mikolov, Tomas and Schuster, Mike and Ge, Qi and Brants, Thorsten and Koehn, Phillipp and Robinson, Tony},
  booktitle={arXiv preprint arXiv:1312.3005},
  year={2013}
}

@article{Ding2023,
  doi = {10.1038/s42256-023-00626-4},
  url = {https://doi.org/10.1038/s42256-023-00626-4},
  year = {2023},
  month = mar,
  publisher = {Springer Science and Business Media {LLC}},
  volume = {5},
  number = {3},
  pages = {220--235},
  author = {Ning Ding and Yujia Qin and Guang Yang and Fuchao Wei and Zonghan Yang and Yusheng Su and Shengding Hu and Yulin Chen and Chi-Min Chan and Weize Chen and Jing Yi and Weilin Zhao and Xiaozhi Wang and Zhiyuan Liu and Hai-Tao Zheng and Jianfei Chen and Yang Liu and Jie Tang and Juanzi Li and Maosong Sun},
  title = {Parameter-efficient fine-tuning of large-scale pre-trained language models},
  journal = {Nature Machine Intelligence}
}

@inproceedings{su-etal-2022-transferability,
    title = "On Transferability of Prompt Tuning for Natural Language Processing",
    author = "Su, Yusheng  and
      Wang, Xiaozhi  and
      Qin, Yujia  and
      Chan, Chi-Min  and
      Lin, Yankai  and
      Wang, Huadong  and
      Wen, Kaiyue  and
      Liu, Zhiyuan  and
      Li, Peng  and
      Li, Juanzi  and
      Hou, Lei  and
      Sun, Maosong  and
      Zhou, Jie",
    booktitle = "Proceedings of the 2022 Conference of the North American Chapter of the Association for Computational Linguistics: Human Language Technologies",
    month = jul,
    year = "2022",
    address = "Seattle, United States",
    publisher = "Association for Computational Linguistics",
    url = "https://aclanthology.org/2022.naacl-main.290",
    doi = "10.18653/v1/2022.naacl-main.290",
    pages = "3949--3969",
}

@inproceedings{vu-etal-2022-spot,
    title = "{SP}o{T}: Better Frozen Model Adaptation through Soft Prompt Transfer",
    author = "Vu, Tu  and
      Lester, Brian  and
      Constant, Noah  and
      Al-Rfou{'}, Rami  and
      Cer, Daniel",
    booktitle = "Proceedings of the 60th Annual Meeting of the Association for Computational Linguistics (Volume 1: Long Papers)",
    month = may,
    year = "2022",
    address = "Dublin, Ireland",
    publisher = "Association for Computational Linguistics",
    url = "https://aclanthology.org/2022.acl-long.346",
    doi = "10.18653/v1/2022.acl-long.346",
    pages = "5039--5059",
}

\appendix

\begin{table*}[h]
    \centering
    \small
    \setlength\tabcolsep{2pt} % default value: 6pt
    \renewcommand{\arraystretch}{1.10}
    \begin{tabular}{|ll|ccc|ccc|}
    \hline
        \multirow{2}{*}{\textbf{Dataset}} & \multirow{2}{*}{\textbf{Split}} & \multicolumn{3}{c|}{\textbf{Before sampling}} & \multicolumn{3}{c|}{\textbf{After sampling}} \\ \cline{3-8}
         & & \textbf{Num samples} & \textbf{Avg words} & \textbf{Avg char} & \textbf{Num samples} & \textbf{Avg words} & \textbf{Avg char} \\
         \hline
         \multirow{2}{*}{IMDb} & train & 25000 & 231.49 & 1302.62 & 24330 & 231.50 & 1302.54 \\
          & test & 25000 & 226.25 & 1243.82 & 100 & 218.61 & 1271.64 \\
         \hline
         \multirow{3}{*}{SST-2} & train & 67349 & 9.41 & 97.21 & 24330 & 17.86 & 53.03 \\
          & validation & 872 & 19.55 & 116.24 & 726 & 21.62 & 105.05 \\
           & test & 1821 & 19.23 & 49.86 & 100 & 9.33 & 103.15 \\
         \hline
         \multirow{3}{*}{Yelp} & train & 470000 & 134.03 & 728.80 & 24330 & 135.85 & 718.88 \\
          & validation & 50000 & 133.87 & 746.85 & 726 & 138.81 & 717.61 \\
           & test & 40000 & 133.87 & 712.02 & 100 & 132.03 & 718.08 \\
         \hline
           \multirow{2}{*}{Combined Sentiment} & train & - & - & - & 72990 & 128.41 & 709.54 \\
            & validation & - & - & - & 1452 & 80.22 & 431.55  \\
            \hline \hline
         \multirow{3}{*}{AG News} & train & 112000 & 37.85 & 236.11 & 111992 & 37.86 & 236.10 \\
          & validation & 8000 & 37.78 & 235.75 & 3996 & 37.78 & 235.67 \\
           & test & 7600 & 37.72 & 231.37 & 200 & 37.42 & 234.90 \\
         \hline
         \multirow{3}{*}{DBpedia} & train & 546000 & 46.13 & 275.05 & 111984 & 44.91 & 280.56 \\
          & validation & 14000 & 46.16 & 274.94 & 3996 & 44.90 & 280.49 \\
           & test & 70000 & 46.14 & 265.66 & 195 & 43.12 & 280.57 \\
           \hline
           \multirow{2}{*}{Combined Topic} & train & - & - & - & 223976 & 41.38 & 255.58 \\
            & validation & - & - & - & 7992 & 41.34 & 255.35 \\
           \hline \hline
           PPLM & test & 35 & 2.54 & 13.31 & 35 & 2.54 & 13.31 \\
           STS & test & 625 & 8.02 & 39.36 & 100 & 7.50 & 36.83 \\
           STS proc & test & - & - & - & 100 & 3.18 & 13.71 \\
         \hline
    \end{tabular}
    \caption{Statistics of the datasets used for each data split (train, validation, test), showing the number of samples, average number of words per text, and average number of characters per text, both before and after dataset sampling.}
    \label{tab:dataset_descr}
\end{table*}

\begin{table}[]
    \centering
    \begin{tabular}{|l|l|}
        \hline
        \textbf{Original label} & \textbf{Mapped label} \\
        \hline
        Company	& Business \\
        Educational & World \\
        Artist & World \\
        Athlete	& Sports \\
        OfficeHolder & Business \\
        MeanOfTransportation & World \\
        Building & World \\
        NaturalPlace & World \\
        Village & World \\
        Animal & Sci/Tech \\
        Plant & Sci/Tech \\
        Album & World \\
        Film & World \\
        WrittenWork & World \\
        \hline
    \end{tabular}
    \caption{Mapping of DBPedia topic labels into AG News topic labels}
    \label{tab:dbpedia_map}
\end{table}

\begin{figure}
    \centering
    \includegraphics[width=0.8\linewidth]{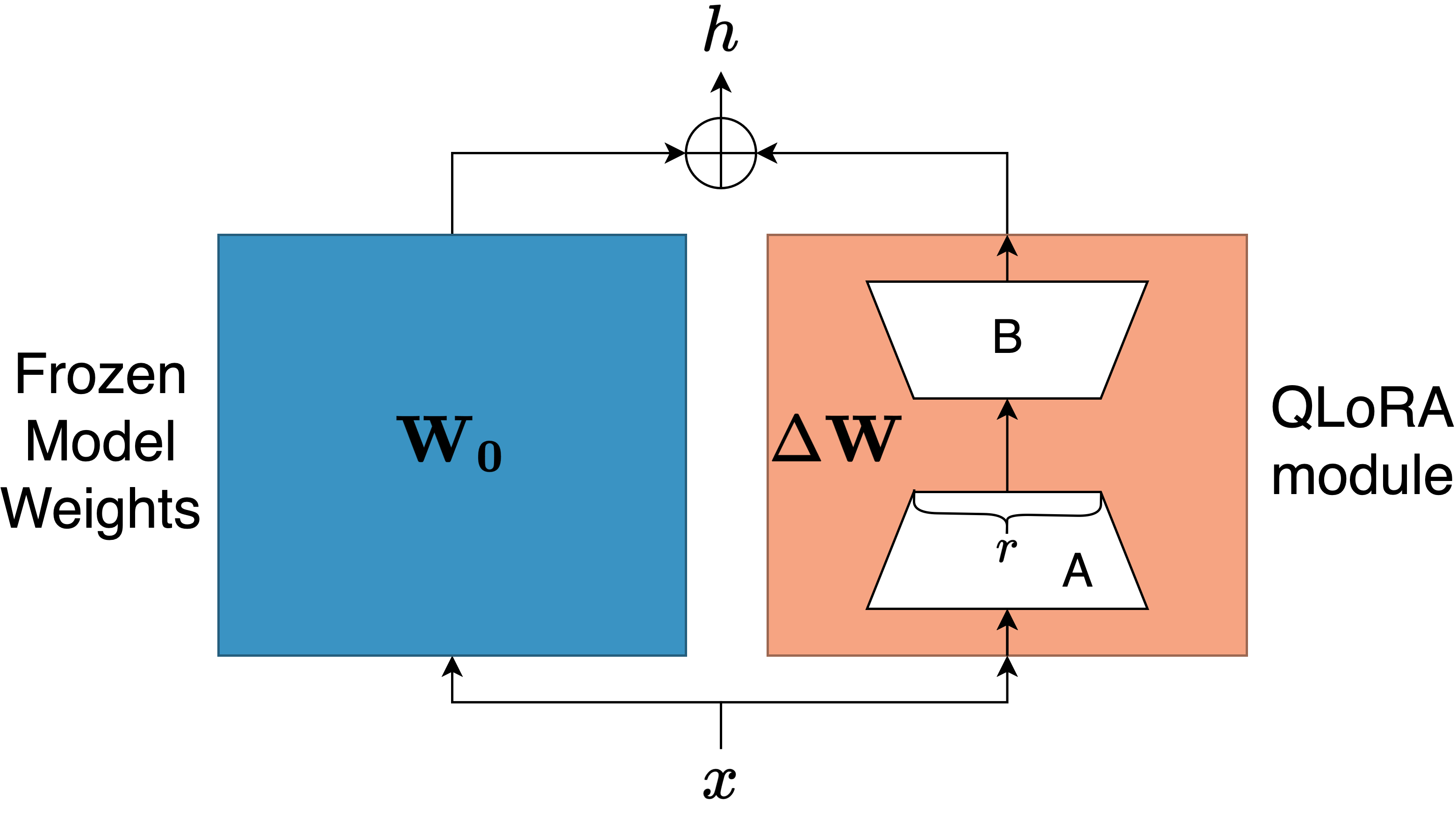}
    \caption{A single QLoRA block (orange) shown attached to its corresponding model weights. }
    \label{fig:qlora}
\end{figure}

\section{Composite dataset construction}\label{sec:sampling}

As a baseline point of comparison we finetune QLoRA modules on a composite dataset made up of multiple (subsets of)  datasets from Section~\ref{sec:model_data}, created via the following three-step process:
\begin{enumerate}%\itemsep=-.01cm
\item \textbf{Filtering}: Dataset items with fewer than 10 words, and any neutral-tagged sentiment dataset items (just Yelp) are removed. 
\item \textbf{Smallest dataset balancing}: We identify the smallest dataset; if its label distribution is imbalanced, it is randomly downsampled to get a balanced distribution of items over labels. 
\item \textbf{Stratified sampling}: All other datasets are then downsampled to match the size of the smallest dataset, ensuring that each dataset is also balanced across labels and text length. 
\end{enumerate}

\noindent We combine the three sampled sentiment datasets to form the Combined Sentiment dataset, and the two topic datasets to form the Combined Topic dataset. For these combined datasets, we sample only from the training and validation splits. In contrast, for the single-dataset test sets, we include up to 50 examples per label. For descriptive statistics of datasets see Appendix~\ref{app:data}.

\section{Dataset Descriptives}
\label{app:data}

Table~\ref{tab:dataset_descr} shows a detailed overview of the datasets used, including details of each data split (train, validation, and test). The table reports the number of samples, the average number of words per text, and the average number of characters per text before and after stratified sampling. We can notice that the stratified sampling based on text length effectively created balanced datasets while preserving the diversity of input length.

As discussed in Section~\ref{sec:model_data}, it is necessary to align the topic labels in the DBPedia dataset with the topic labels in the AG News dataset to ensure a consistent set of topics across both datasets. Table~\ref{tab:dbpedia_map} shows the mapping from the original DBPedia topics to the AG News topics.

\begin{table*}[]
    \centering
    \scriptsize
    \setlength\tabcolsep{2pt} % default value: 6pt
    \renewcommand{\arraystretch}{1.10}
    \begin{tabular}{|l|c|ccc|cccc|}
    \hline
         \multirow{3}{*}{\textbf{CTG Technique}} & \multicolumn{8}{c|}{\textbf{Control Effectiveness}$\uparrow$} \\ \cline{2-9}
         & \multirow{2}{*}{\textbf{Avg All}} & \multirow{2}{*}{\textbf{Avg}} & \multirow{2}{*}{\textbf{AG News}} & \multirow{2}{*}{\textbf{DBPedia}} & \multicolumn{4}{c|}{\textbf{Out-Of-Domain}} \\ \cline{6-9}
         &  &  &  &  & \textbf{Avg} & \textbf{PPLM T} & \textbf{STS T} & \textbf{STS proc T} \\
\hline
Llama 3 8B & 54.49 & 73.22 & 75.67 & 70.77 & 43.25 & 60.24 & 27.92 & 41.58 \\
+ QLoRA AG News & \textbf{68.82} & \textbf{84.56} & \textbf{84.50} & \textbf{84.62} & \textbf{59.72} & \textbf{75.24} & \textbf{42.67} & \cellcolor{gray!25}61.25 \\
+ QLoRA DBPedia & 59.43 & 78.31 & 78.83 & 77.78 & 48.21 & 65.71 & 31.17 & 47.75 \\
+ QLoRA Combined Topic dataset & 63.51 & 75.19 & 75.50 & 74.87 & \cellcolor{gray!25}58.22 & \cellcolor{gray!25}75.00 & \cellcolor{gray!25}37.00 & \textbf{62.67} \\
\hdashline
+ QLoRA Output Summing(AG News, DBPedia) & \cellcolor{gray!25}64.44 & \cellcolor{gray!25}83.02 & \cellcolor{gray!25}84.33 & \cellcolor{gray!25}81.71 & 53.87 & 72.86 & 35.17 & 53.58 \\
+ QLoRA Output Averaging(AG News, DBPedia) & 63.05 & 81.60 & 81.67 & 81.54 & 52.76 & 72.62 & 32.83 & 52.83 \\
+ QLoRA Averaged Weights(AG News, DBPedia) & 60.64 & 79.31 & 80.50 & 78.12 & 50.58 & 71.67 & 31.25 & 48.83 \\
\hline \hline
Llama 3.1 8B & 53.08 & 68.65 & 71.83 & 65.47 & 44.81 & 62.86 & 32.92 & 38.67 \\
+ QLoRA AG News & \textbf{68.09} & \textbf{84.38} & \textbf{85.00} & \cellcolor{gray!25}83.76 & \textbf{58.41} & \textbf{73.57} & \cellcolor{gray!25}36.92 & \textbf{64.75} \\
+ QLoRA DBPedia & 59.17 & 76.73 & 75.17 & 78.29 & 49.34 & 67.86 & 32.42 & 47.75 \\
+ QLoRA Combined Topic dataset & \cellcolor{gray!25}66.31 & 82.60 & \cellcolor{gray!25}83.50 & 81.71 & \cellcolor{gray!25}56.35 & 70.48 & \textbf{39.17} & \cellcolor{gray!25}59.42 \\
\hdashline
+ QLoRA Output Summing(AG News, DBPedia) & 64.68 & \cellcolor{gray!25}83.64 & 82.83 & \textbf{84.44} & 53.45 & \cellcolor{gray!25}71.19 & 30.50 & 58.67 \\
+ QLoRA Output Averaging(AG News, DBPedia) & 62.43 & 81.17 & 82.00 & 80.34 & 51.48 & 69.52 & 30.08 & 54.83 \\
+ QLoRA Averaged Weights(AG News, DBPedia) & 58.20 & 75.89 & 74.17 & 77.61 & 48.51 & 67.86 & 30.92 & 46.75 \\
\hline \hline
Mistral 7B & 49.63 & 65.81 & 66.50 & 65.13 & 42.48 & 66.43 & 29.00 & 32.00 \\
+ QLoRA AG News & \textbf{69.06} & \textbf{88.63} & \textbf{87.00} & \textbf{90.26} & \textbf{56.89} & \cellcolor{gray!25}73.10 & \textbf{40.75} & \textbf{56.83} \\
+ QLoRA DBPedia & 59.00 & 72.41 & 72.00 & 72.82 & 51.67 & 66.43 & 35.33 & 53.25 \\
+ QLoRA Combined Topic dataset & 65.04 & 82.62 & 82.33 & 82.91 & 54.39 & 70.00 & 37.08 & \cellcolor{gray!25}56.08 \\
\hdashline
+ QLoRA Output Summing(AG News, DBPedia) & \cellcolor{gray!25}66.01 & \cellcolor{gray!25}83.98 & \cellcolor{gray!25}83.00 & \cellcolor{gray!25}84.96 & \cellcolor{gray!25}56.08 & \textbf{75.48} & \cellcolor{gray!25}39.75 & 53.00 \\
+ QLoRA Output Averaging(AG News, DBPedia) & 60.81 & 80.34 & 80.17 & 80.51 & 49.73 & 69.76 & 33.00 & 46.42 \\
+ QLoRA Averaged Weights(AG News, DBPedia) & 59.13 & 72.41 & 72.17 & 72.65 & 52.23 & 68.10 & 35.17 & 53.42 \\
\hline
    \end{tabular}
    \caption{Control Effectiveness calculated using an LLM as classifier for \textbf{Topic Control}, training on \textit{single} and \textit{combined} datasets, and composition of modules trained on single datasets Sum(data1, data2), Average(data1, data2), and Weights Average(data1, data2). Bold (shaded) = (second) highest score in column and section; underline = train and test set from same dataset.}
    \label{tab:topic_res_llm}
\end{table*}

\section{Weight Averaging vs Output Averaging}
\label{app:weight_vs_output}

While both weight averaging and output averaging involve linear operations, they are \textit{not} mathematically equivalent due to how the averaging interacts with the low-rank structure. An important distinction between weight averaging and output composition methods arises from our implementation: weight averaging averages the low-rank factors A and B separately ($A_{\text{avg}} = \frac{1}{N} \sum A_i$, $B_{\text{avg}} = \frac{1}{N} \sum B_i$), then computes the adapted output as $A_{\text{avg}} @ B_{\text{avg}}$. This differs fundamentally from output averaging, which computes each module's full transformation $A_i@B_i$ before averaging.

When factors are averaged separately, the resulting computation includes cross-terms: 
\[
[\sum A_i]@[\sum B_j] = \sum_{i}(A_i B_i) + \sum_{i\neq j}(A_i B_j).
\]
The cross-terms $A_i B_j$ ($i\neq j$) represent combinations of the down-projection from module i with the up-projection from module j-components that were never trained to work together. These cross-terms are absent in output averaging, which only combines the trained transformations $A_i B_i$. For N modules, weight averaging produces $N^{2}$ terms while output averaging produces N terms. Additionally, averaging factors separately results in $1/N^{2}$ scaling, compared to $1/N$ for output averaging (or no normalization for output summing).

The presence of cross-terms means that weight averaging creates novel, untrained combinations of module components, which may explain why output composition methods consistently outperform weight averaging in our experiments (Tables~\ref{tab:sent_res}-\ref{tab:multi_res}).

\section{Models}\label{app:models}
In this study, we consider three different raw pretrained LLMs: LLaMa 3 8B~\citep{llama3modelcard}, LLaMa 3.1 8B, and Mistral 7B~\citep{jiang2023mistral}.

LLaMa 3 8B is a pre-trained autoregressive transformer model with 8 billion parameters. It has a decoder-only architecture and it is optimised with grouped-query attention for faster inference and SwiGLU activation function for improved training efficiency. Additionally, it incorporates Rotary Positional Embeddings (RoPE) to handle longer context lengths. The model was trained on 15T tokens that were collected from publicly available sources and highly curated to get a large high-quality training dataset.\footnote{\url{https://ai.meta.com/blog/meta-llama-3/}}

LLaMA 3.1 8B is an incremental update to LLaMA 3 8B, with the same 8 billion parameter transformer architecture but incorporating improvements in training stability, dataset quality, and tokenisation. While specific architectural details are not publicly disclosed, this version is expected to refine pretraining efficiency, dataset diversity, and representation quality, leading to better perplexity scores and more coherent text generation.

Mistral 7B is a pretrained transformer model with 7 billion parameters, designed to be highly efficient while maintaining strong language modeling performance. It incorporates grouped-query attention together with sliding window attention, which allows to efficiently handle longer context lengths without increasing computational complexity.

\begin{table}[]
    \centering
    \begin{tabular}{|l|cc|}
        \hline
        \textbf{Dataset} & \textbf{Pearson} & \textbf{Spearman} \\ \hline
        AG News & 0.915 (0.000) & 0.944 (0.000) \\
DBPedia & 0.955 (0.000) & 0.989 (0.000) \\
PPLM Prompt & 0.917 (0.000) & 0.928 (0.000) \\
STS & 0.919 (0.000) & 0.883 (0.000) \\
STS proc & 0.941 (0.000) & 0.944 (0.000) \\
\hline
    \end{tabular}
    \caption{Pearson's and Spearman's correlations between the LLM as classifier and the other three classifiers used. p-value in parentheses.}
    \label{tab:corr_llm}
\end{table}

\section{Comparison of CE with LLM as classifier}
\label{app:llm_classifier}

\begin{table*}[]
    \centering
    \begin{tabular}{|>{\raggedright\arraybackslash}p{0.4\textwidth}|>{\raggedright\arraybackslash}p{0.6\textwidth}|}
        \hline
         \textbf{System message:} & You are a helpful assistant that classifies texts by topic. \\ \hline
         \textbf{User message:} & Classify the following text into one of these topic categories: SPORTS, BUSINESS, WORLD, SCIENCE/TECHNOLOGY.\par Only reply with one of the possible topic categories. Do not include any other category or text.\bigbreak \{text\} \\
         \hline
    \end{tabular}
    \caption{Prompt used for the LLM as classifier.}
    \label{tab:prompt_llm}
\end{table*}

Given that all our classifiers are trained on AG News, our evaluation setup might be biased favouring the texts generated by a system trained on AG News. To test whether this is the case or not, we performed an additional evaluation using a general-purpose instruction-tuned LLM as a classifier. We used Command R plus quantised in 4 bit,\footnote{\url{https://huggingface.co/CohereForAI/c4ai-command-r-plus-4bit}} using a simple prompt (Table~\ref{tab:prompt_llm}) following Cohere's guidelines and template for classification.\footnote{\url{https://cohere.com/llmu/use-case-patterns\#classifying}}$^,$\footnote{\url{https://huggingface.co/CohereForAI/c4ai-command-r-plus-4bit}} The designed prompt is then formatted with the correct special tokens using the apply chat template function.

Our findings show that the overall trends, i.e.\ the settings identified as best or second-best, remain consistent with those observed in Table~\ref{tab:topic_res}, indicating that the classifiers’ training did not significantly distort our conclusions. To further validate this hypothesis, we computed Pearson's and Spearman's correlation coefficients between the Control Effectiveness scores obtained from the instruction-tuned LLM and those from the original classifiers.

The results (Table~\ref{tab:corr_llm}) show strong correlations, with Pearson's correlation ranging from 0.915 for AG News to 0.955 for DBPedia, and Spearman's correlation ranging from 0.883 for STS to 0.944 for AG News and STS proc. These high correlation values confirm that the instruction-tuned LLM behaves similarly to the classifiers originally used in our evaluation, reinforcing the reliability of our Control Effectiveness metric.

\section{Module Composition Analysis}
\label{sec:app_mod_comp}

For a closer look at module composition not only preserving single-task performance, but even outperforming task-specialised single modules, Table~\ref{tab:out_sum} shows performance of models tested on individual test data sets ($D_a$) when trained on (just) $D_a$, compared to 2-module and 3-module compositions, for the sentiment control task.

\begin{table}[h!]
    \centering
    \small
     \setlength\tabcolsep{2.5pt} % default value: 6pt
   \begin{tabular}{|l|l|c|c|c|c|}
\hline
        & test & \multicolumn{4}{c|}{training / output summing } \\
        \cline{2-6}
Model   & $D_a$  & $D_a$ & $D_a,D_b$ & $D_a,D_c$ & $D_a,D_b,D_c$ \\
\hline
        & IMDB & 89.67         & 89.33          & 89.56          & \textbf{92.56} \\
L3-8B   & SST2 & \cellcolor{gray!25}91.33& 89.56          & \textbf{93.22} & 90.44 \\
        & Yelp & 92.78         & 90.22          & 92.44          & \textbf{94.44} \\
\hline
        & IMDB & 91.44         & 90.00          & 90.89          & \textbf{92.67} \\
L3.1-8B & SST2 & 91.22         & 93.22          & 94.44          & \textbf{95.11} \\
        & Yelp & 91.00         & 91.22          & 92.78          & \textbf{96.33} \\
\hline
        & IMDB & 88.56         & 85.44          & 86.33          & \textbf{89.44} \\
Mist7B  & SST2 & 92.22         & 90.33          & 93.44          & \textbf{95.22} \\
        & Yelp & \textbf{94.78}& 89.00          & 93.33          & \cellcolor{gray!25}93.78 \\
\hline  
\multicolumn{2}{|c|}{Average}& 91.44	       & 89.81	        & \cellcolor{gray!25}91.83	         & \textbf{93.33} \\
\hline
\end{tabular}
\caption{
Sentiment control effectiveness (\%) comparing single-dataset training with multi-module output summing. Bold=best; gray=second-best.
}
\label{tab:out_sum}
\end{table}

The very clear trend is that the 3-module composition outperforms the others in 7 of 9 scenarios (average 93.33\% vs. 91.44\% for single-task modules), demonstrating (i) generalisation to the sentiment-control task generally (beyond individual datasets), and (ii) an advantage for single-dataset performance resulting from the generalisation. The two exceptions, L3-8B on SST-2 (-0.89\%) and Mistral-7B on Yelp (-1.00\%), represent marginal differences that do not change the core finding: output composition consistently matches or exceeds specialised module performance while enabling plug-and-play multi-task functionality.

\section{Standard Deviation Results}
\label{app:res_std}

In all tables we report the results averaged across three runs with different seeds. We report all the results including standard deviation in Tables~\ref{tab:sent_res_std_dist} and \ref{tab:sent_res_std} for sentiment control, in Tables~\ref{tab:topic_res_std_dist} and \ref{tab:topic_res_std_ce} for topic control, and in Tables~\ref{tab:multi_res_std_dist} and \ref{tab:multi_res_std_ce} for multiple-attribute control.

\begin{table*}[h!]
    \centering
    \scriptsize
    \setlength\tabcolsep{2pt} % default value: 6pt
    \renewcommand{\arraystretch}{1.10}
    \begin{tabular}{|l|ccc|c|}
    \hline
         \multirow{2}{*}{\textbf{CTG Technique}} & \multicolumn{3}{c|}{\textbf{Distinct-n}$\uparrow$} & \multirow{2}{*}{\textbf{SLOR}$\uparrow$} \\ \cline{2-4}
         & \textbf{dist-1} & \textbf{dist-2} & \textbf{dist-3} &  \\
\hline \hline
Llama 3 8B & 0.03$\pm$ 0.01 & 0.09$\pm$ 0.02 & 0.12$\pm$ 0.03 & 9.44$\pm$ 0.59 \\
+ QLoRA Yelp & 0.09$\pm$ 0.04 & 0.24$\pm$ 0.11 & 0.34$\pm$ 0.15 & 9.87$\pm$ 0.40 \\
+ QLoRA IMDB & 0.04$\pm$ 0.02 & 0.13$\pm$ 0.04 & 0.19$\pm$ 0.06 & \textbf{10.86}$\pm$ 0.30 \\
+ QLoRA SST-2 & \textbf{0.34}$\pm$ 0.08 & \textbf{0.64}$\pm$ 0.12 & \textbf{0.71}$\pm$ 0.12 & 8.24$\pm$ 0.12 \\
+ QLoRA Combined Sentiment dataset & 0.11$\pm$ 0.10 & 0.27$\pm$ 0.20 & 0.35$\pm$ 0.22 & 10.07$\pm$ 0.95 \\
\hdashline
+ QLoRA Output Summing(IMDB, SST-2) & 0.15$\pm$ 0.09 & 0.36$\pm$ 0.19 & 0.46$\pm$ 0.22 & 9.43$\pm$ 0.39 \\
+ QLoRA Output Summing(IMDB, Yelp) & 0.07$\pm$ 0.04 & 0.21$\pm$ 0.09 & 0.31$\pm$ 0.13 & 10.41$\pm$ 0.38 \\
+ QLoRA Output Summing(Yelp, SST-2) & 0.21$\pm$ 0.11 & 0.46$\pm$ 0.22 & 0.56$\pm$ 0.24 & 8.93$\pm$ 0.34 \\
+ QLoRA Output Summing(IMDB, Yelp, SST-2) & 0.13$\pm$ 0.07 & 0.34$\pm$ 0.18 & 0.45$\pm$ 0.23 & 9.83$\pm$ 0.50 \\
\dottedline
+ QLoRA Output Averaging(IMDB, SST-2) & 0.12$\pm$ 0.07 & 0.27$\pm$ 0.14 & 0.35$\pm$ 0.16 & 9.49$\pm$ 0.37 \\
+ QLoRA Output Averaging(IMDB, Yelp) & 0.05$\pm$ 0.02 & 0.16$\pm$ 0.06 & 0.23$\pm$ 0.08 & 10.30$\pm$ 0.25\\
+ QLoRA Output Averaging(Yelp, SST-2) & 0.13$\pm$ 0.08 & 0.31$\pm$ 0.16 & 0.38$\pm$ 0.19 & 9.16$\pm$ 0.41 \\
+ QLoRA Output Averaging(IMDB, Yelp, SST-2) & 0.07$\pm$ 0.04 & 0.19$\pm$ 0.09 & 0.27$\pm$ 0.12 & 9.77$\pm$ 0.39 \\
\dottedline
+ QLoRA Averaged Weights(IMDB, SST-2) & 0.31$\pm$ 0.11 & \cellcolor{gray!25}0.61$\pm$ 0.18 & \cellcolor{gray!25}0.69$\pm$ 0.17 & 8.37$\pm$ 0.50 \\
+ QLoRA Averaged Weights(IMDB, Yelp) & 0.04$\pm$ 0.02 & 0.12$\pm$ 0.05 & 0.19$\pm$ 0.06 & \cellcolor{gray!25}10.83$\pm$ 0.34 \\
+ QLoRA Averaged Weights(Yelp, SST-2) & \cellcolor{gray!25}0.33$\pm$ 0.08 & \textbf{0.64}$\pm$ 0.13 & \textbf{0.71}$\pm$ 0.12 & 8.28$\pm$ 0.07 \\
+ QLoRA Averaged Weights(IMDB, Yelp, SST-2) & 0.11$\pm$ 0.06 & 0.28$\pm$ 0.14 & 0.36$\pm$ 0.17 & 9.54$\pm$ 0.40 \\
\hline \hline
Llama 3.1 8B & 0.04$\pm$ 0.01 & 0.10$\pm$ 0.03 & 0.13$\pm$ 0.04 & 9.66$\pm$ 0.63 \\
+ QLoRA Yelp & 0.09$\pm$ 0.05 & 0.26$\pm$ 0.13 & 0.36$\pm$ 0.17 & 9.94$\pm$ 0.50 \\
+ QLoRA IMDB & 0.04$\pm$ 0.02 & 0.14$\pm$ 0.05 & 0.21$\pm$ 0.07 & \textbf{10.81}$\pm$ 0.33 \\
+ QLoRA SST-2 & \textbf{0.36}$\pm$ 0.07 & \textbf{0.69}$\pm$ 0.11 & \textbf{0.76}$\pm$ 0.11 & 8.24$\pm$ 0.11 \\
+ QLoRA Combined Sentiment dataset & 0.12$\pm$ 0.11 & 0.27$\pm$ 0.19 & 0.35$\pm$ 0.21 & 10.00$\pm$ 0.96 \\
\hdashline
+ QLoRA Output Summing(IMDB, SST-2) & 0.09$\pm$ 0.04 & 0.23$\pm$ 0.11 & 0.31$\pm$ 0.14 & 9.91$\pm$ 0.49 \\
+ QLoRA Output Summing(IMDB, Yelp) & 0.07$\pm$ 0.04 & 0.21$\pm$ 0.11 & 0.31$\pm$ 0.16 & 10.33$\pm$ 0.56 \\
+ QLoRA Output Summing(Yelp, SST-2) & 0.16$\pm$ 0.09 & 0.38$\pm$ 0.19 & 0.47$\pm$ 0.23 & 9.16$\pm$ 0.74 \\
+ QLoRA Output Summing(IMDB, Yelp, SST-2) & 0.10$\pm$ 0.06 & 0.27$\pm$ 0.14 & 0.37$\pm$ 0.19 & 9.91$\pm$ 0.76 \\
\dottedline
+ QLoRA Output Averaging(IMDB, SST-2) & 0.08$\pm$ 0.04 & 0.21$\pm$ 0.09 & 0.28$\pm$ 0.11 & 9.81$\pm$ 0.58 \\
+ QLoRA Output Averaging(IMDB, Yelp) & 0.07$\pm$ 0.04 & 0.22$\pm$ 0.10 & 0.31$\pm$ 0.14 & 10.24$\pm$ 0.34 \\
+ QLoRA Output Averaging(Yelp, SST-2) & 0.16$\pm$ 0.10 & 0.36$\pm$ 0.20 & 0.44$\pm$ 0.23 & 9.05$\pm$ 0.59 \\
+ QLoRA Output Averaging(IMDB, Yelp, SST-2) & 0.07$\pm$ 0.04 & 0.18$\pm$ 0.08 & 0.26$\pm$ 0.11 & 10.01$\pm$ 0.60 \\
\dottedline
+ QLoRA Averaged Weights(IMDB, SST-2) & 0.33$\pm$ 0.12 & 0.64$\pm$ 0.21 & 0.71$\pm$ 0.20 & 8.38$\pm$ 0.61 \\
+ QLoRA Averaged Weights(IMDB, Yelp) & 0.04$\pm$ 0.03 & 0.14$\pm$ 0.06 & 0.21$\pm$ 0.08 & \cellcolor{gray!25}10.78$\pm$ 0.42 \\
+ QLoRA Averaged Weights(Yelp, SST-2) & \cellcolor{gray!25}0.35$\pm$ 0.08 & \cellcolor{gray!25}0.67$\pm$ 0.12 & \cellcolor{gray!25}0.74$\pm$ 0.11 & 8.27$\pm$ 0.12 \\
+ QLoRA Averaged Weights(IMDB, Yelp, SST-2) & 0.08$\pm$ 0.04 & 0.20$\pm$ 0.09 & 0.27$\pm$ 0.12 & 9.87$\pm$ 0.59 \\
\hline \hline
Mistral 7B & 0.04$\pm$ 0.02 & 0.09$\pm$ 0.03 & 0.12$\pm$ 0.03 & 7.45$\pm$ 1.72 \\
+ QLoRA Yelp & 0.03$\pm$ 0.01 & 0.07$\pm$ 0.03 & 0.10$\pm$ 0.04 & 10.44$\pm$ 0.68 \\
+ QLoRA IMDB & 0.02$\pm$ 0.01 & 0.04$\pm$ 0.02 & 0.06$\pm$ 0.03 & \textbf{11.25}$\pm$ 0.95 \\
+ QLoRA SST-2 & \textbf{0.27}$\pm$ 0.06 & \textbf{0.51}$\pm$ 0.10 & \textbf{0.58}$\pm$ 0.10 & 7.65$\pm$ 0.19 \\
+ QLoRA Combined Sentiment dataset & 0.03$\pm$ 0.03 & 0.06$\pm$ 0.05 & 0.09$\pm$ 0.05 & 10.83$\pm$ 1.13 \\
\hdashline
+ QLoRA Output Summing(IMDB, SST-2) & 0.22$\pm$ 0.09 & 0.41$\pm$ 0.15 & 0.47$\pm$ 0.15 & 7.80$\pm$ 0.47 \\
+ QLoRA Output Summing(IMDB, Yelp) & 0.03$\pm$ 0.01 & 0.07$\pm$ 0.03 & 0.10$\pm$ 0.04 & 10.51$\pm$ 0.68 \\
+ QLoRA Output Summing(Yelp, SST-2) & 0.18$\pm$ 0.10 & 0.35$\pm$ 0.17 & 0.40$\pm$ 0.20 & 8.22$\pm$ 1.05 \\
+ QLoRA Output Summing(IMDB, Yelp, SST-2) & 0.08$\pm$ 0.08 & 0.16$\pm$ 0.13 & 0.19$\pm$ 0.13 & 9.13$\pm$ 1.12 \\
\dottedline
+ QLoRA Output Averaging(IMDB, SST-2) & 0.11$\pm$ 0.07 & 0.21$\pm$ 0.11 & 0.25$\pm$ 0.12 & 8.49$\pm$ 0.56 \\
+ QLoRA Output Averaging(IMDB, Yelp) & 0.02$\pm$ 0.01 & 0.06$\pm$ 0.03 & 0.09$\pm$ 0.04 & 10.80$\pm$ 0.72 \\
+ QLoRA Output Averaging(Yelp, SST-2) & 0.05$\pm$ 0.03 & 0.10$\pm$ 0.06 & 0.13$\pm$ 0.06 & 8.84$\pm$ 0.61 \\
+ QLoRA Output Averaging(IMDB, Yelp, SST-2) & 0.02$\pm$ 0.01 & 0.05$\pm$ 0.02 & 0.07$\pm$ 0.02 & 10.54$\pm$ 0.75 \\
\dottedline
+ QLoRA Averaged Weights(IMDB, SST-2) & \cellcolor{gray!25}0.25$\pm$ 0.06 & 0.49$\pm$ 0.11 & 0.55$\pm$ 0.11 & 7.69$\pm$ 0.25 \\
+ QLoRA Averaged Weights(IMDB, Yelp) & 0.02$\pm$ 0.01 & 0.04$\pm$ 0.02 & 0.07$\pm$ 0.03 & \cellcolor{gray!25}11.19$\pm$ 0.93 \\
+ QLoRA Averaged Weights(Yelp, SST-2) & \textbf{0.27}$\pm$ 0.05 & \cellcolor{gray!25}0.50$\pm$ 0.08 & \cellcolor{gray!25}0.57$\pm$ 0.09 & 7.66$\pm$ 0.19 \\
+ QLoRA Averaged Weights(IMDB, Yelp, SST-2) & 0.10$\pm$ 0.07 & 0.20$\pm$ 0.12 & 0.24$\pm$ 0.12 & 8.64$\pm$ 1.07 \\
\hline
    \end{tabular}
    \caption{\textbf{Sentiment Control} Diversity, Fluency for the model + QLoRA module combinations explained in Section~\ref{sec:study_overview}. Here, e.g.\ Output Summing(data1, data2) refers to the output summation module composition technique. All values are averages over 3 runs and standard deviation is reported.
    Bold (shaded) =  (second) highest score in column/section.
    }
    \label{tab:sent_res_std_dist}
\end{table*}

\begin{table*}[h!]
    \centering
    \scriptsize
    \setlength\tabcolsep{2pt} % default value: 6pt
    \renewcommand{\arraystretch}{1.10}
    \begin{tabular}{|l|c|cccc|cccc|}
    \hline
         \multirow{3}{*}{\textbf{CTG Technique}} & \multicolumn{9}{c|}{\textbf{Control Effectiveness}$\uparrow$} \\ \cline{2-10}
         & \multirow{2}{*}{\textbf{Avg All}} & \multirow{2}{*}{\textbf{Avg}} & \multirow{2}{*}{\textbf{Yelp}} & \multirow{2}{*}{\textbf{IMDB}} & \multirow{2}{*}{\textbf{SST-2}} & \multicolumn{4}{c|}{\textbf{Out-Of-Domain}} \\ \cline{7-10}
         &  &  &  &  &  & \textbf{Avg} & \textbf{PPLM S} & \textbf{STS S} & \textbf{STS proc S} \\
\hline \hline
Llama 3 8B & 60.43 & 63.70 & 68.00$\pm$ 2.03 & 57.56$\pm$ 3.37 & 65.56$\pm$ 1.68 & 57.16 & 64.60$\pm$ 3.97 & 52.83$\pm$ 0.88 & 54.06$\pm$ 2.06 \\
+ QLoRA Yelp & \cellcolor{gray!25}85.47 & 91.44 & \cellcolor{gray!25}92.78$\pm$ 1.17 & 90.67$\pm$ 3.48 & 90.89$\pm$ 0.51 & \cellcolor{gray!25}79.49 & 88.41$\pm$ 1.92 & \textbf{67.78}$\pm$ 0.69 & \cellcolor{gray!25}82.28$\pm$ 1.29 \\
+ QLoRA IMDB & 82.84 & 92.07 & 92.11$\pm$ 1.07 & 89.67$\pm$ 1.33 & \cellcolor{gray!25}94.44$\pm$ 0.19 & 73.61 & 80.95$\pm$ 1.65 & 62.00$\pm$ 1.96 & 77.89$\pm$ 2.44 \\
+ QLoRA SST-2 & 78.35 & 85.96 & 84.11$\pm$ 3.47 & 82.44$\pm$ 2.71 & 91.33$\pm$ 2.40 & 70.73 & 82.86$\pm$ 5.85 & 57.39$\pm$ 2.01 & 71.94$\pm$ 1.78 \\
+ QLoRA Combined Sentiment dataset & 83.47 & 91.07 & 92.22$\pm$ 2.41 & \cellcolor{gray!25}92.00$\pm$ 3.18 & 89.00$\pm$ 3.71 & 75.88 & 87.46$\pm$ 2.15 & 63.50$\pm$ 2.19 & 76.67$\pm$ 2.84 \\
\hdashline
+ QLoRA Output Summing(IMDB, SST-2) & 82.85 & 90.85 & 89.78$\pm$ 3.75 & 89.56$\pm$ 0.51 & 93.22$\pm$ 1.64 & 74.85 & \cellcolor{gray!25}88.89$\pm$ 4.16 & 61.72$\pm$ 4.19 & 73.94$\pm$ 2.26 \\
+ QLoRA Output Summing(IMDB, Yelp) & 84.35 & 91.22 & 90.22$\pm$ 0.84 & 89.33$\pm$ 2.03 & 94.11$\pm$ 1.17 & 77.49 & 87.46$\pm$ 1.92 & 64.44$\pm$ 0.98 & 80.56$\pm$ 1.68 \\
+ QLoRA Output Summing(Yelp, SST-2) & 82.85 & 90.74 & 92.44$\pm$ 1.07 & 90.22$\pm$ 2.87 & 89.56$\pm$ 1.68 & 74.96 & \textbf{90.00}$\pm$ 3.12 & 58.33$\pm$ 1.69 & 76.56$\pm$ 3.13 \\
+ QLoRA Output Summing(IMDB, Yelp, SST-2) & \textbf{86.01} & \textbf{92.48} & \textbf{94.44}$\pm$ 1.71 & \textbf{92.56}$\pm$ 0.84 & 90.44$\pm$ 1.71 & \textbf{79.54} & \cellcolor{gray!25}88.89$\pm$ 0.99 & \cellcolor{gray!25}67.00$\pm$ 3.32 & \textbf{82.72}$\pm$ 2.64 \\
\dottedline
+ QLoRA Output Averaging(IMDB, SST-2) & 79.77 & 88.30 & 86.33$\pm$ 0.67 & 87.44$\pm$ 1.02 & 91.11$\pm$ 1.92 & 71.24 & 83.65$\pm$ 3.17 & 58.44$\pm$ 2.34 & 71.61$\pm$ 0.92 \\
+ QLoRA Output Averaging(IMDB, Yelp) & 83.97 & \cellcolor{gray!25}92.26 & 91.11$\pm$ 2.27 & 89.67$\pm$ 4.16 & \textbf{96.00}$\pm$ 0.88 & 75.67 & 86.19$\pm$ 4.59 & 63.17$\pm$ 3.62 & 77.67$\pm$ 4.67 \\
+ QLoRA Output Averaging(Yelp, SST-2) & 82.21 & 89.74 & 89.00$\pm$ 1.73 & 88.56$\pm$ 1.90 & 91.67$\pm$ 2.65 & 74.69 & 84.29$\pm$ 2.90 & 62.61$\pm$ 2.27 & 77.17$\pm$ 1.73 \\
+ QLoRA Output Averaging(IMDB, Yelp, SST-2) & 81.42 & 89.81 & 88.78$\pm$ 0.84 & 89.00$\pm$ 3.06 & 91.67$\pm$ 1.15 & 73.03 & 84.60$\pm$ 2.44 & 60.50$\pm$ 0.83 & 74.00$\pm$ 0.29 \\
\dottedline
+ QLoRA Averaged Weights(IMDB, SST-2) & 77.54 & 85.93 & 85.33$\pm$ 2.08 & 81.89$\pm$ 1.58 & 90.56$\pm$ 0.69 & 69.16 & 77.94$\pm$ 1.80 & 57.50$\pm$ 2.40 & 72.06$\pm$ 2.55 \\
+ QLoRA Averaged Weights(IMDB, Yelp) & 82.20 & 90.74 & 91.89$\pm$ 0.84 & 87.22$\pm$ 2.41 & 93.11$\pm$ 0.69 & 73.65 & 84.13$\pm$ 5.05 & 60.67$\pm$ 1.61 & 76.17$\pm$ 1.17 \\
+ QLoRA Averaged Weights(Yelp, SST-2) & 78.68 & 85.93 & 85.33$\pm$ 2.08 & 81.89$\pm$ 1.58 & 90.56$\pm$ 0.69 & 71.44 & 84.76$\pm$ 1.43 & 57.50$\pm$ 2.40 & 72.06$\pm$ 2.55 \\
+ QLoRA Averaged Weights(IMDB, Yelp, SST-2) & 80.60 & 90.04 & 88.89$\pm$ 2.34 & 88.78$\pm$ 1.90 & 92.44$\pm$ 0.96 & 71.17 & 83.17$\pm$ 2.15 & 58.72$\pm$ 2.08 & 71.61$\pm$ 0.51 \\
\hline \hline
Llama 3.1 8B & 58.96 & 61.52 & 66.22$\pm$ 2.34 & 54.22$\pm$ 1.17 & 64.11$\pm$ 1.35 & 56.41 & 64.29$\pm$ 3.33 & 51.56$\pm$ 2.41 & 53.39$\pm$ 3.51 \\
+ QLoRA Yelp & \cellcolor{gray!25}87.23 & 91.52 & 91.00$\pm$ 2.65 & 89.00$\pm$ 2.65 & 94.56$\pm$ 1.84 & \textbf{82.93} & \cellcolor{gray!25}91.75$\pm$ 3.17 & \textbf{71.28}$\pm$ 1.60 & \textbf{85.78}$\pm$ 1.86 \\
+ QLoRA IMDB & 83.65 & 92.26 & 89.67$\pm$ 3.28 & \cellcolor{gray!25}91.44$\pm$ 1.17 & \textbf{95.67}$\pm$ 1.20 & 75.04 & 85.24$\pm$ 1.26 & 63.33$\pm$ 0.29 & 76.56$\pm$ 3.19 \\
+ QLoRA SST-2 & 78.99 & 85.48 & 82.33$\pm$ 1.45 & 82.89$\pm$ 2.78 & 91.22$\pm$ 1.92 & 72.49 & 84.76$\pm$ 4.54 & 58.17$\pm$ 1.76 & 74.56$\pm$ 0.25 \\
+ QLoRA Combined Sentiment dataset & 85.26 & 92.67 & \cellcolor{gray!25}94.89$\pm$ 2.04 & \cellcolor{gray!25}91.44$\pm$ 0.51 & 91.67$\pm$ 1.20 & 77.86 & 89.52$\pm$ 4.59 & 62.28$\pm$ 1.42 & 81.78$\pm$ 3.14 \\
\hdashline
+ QLoRA Output Summing(IMDB, SST-2) & 85.80 & \cellcolor{gray!25}92.93 & 93.44$\pm$ 1.35 & 90.89$\pm$ 4.55 & 94.44$\pm$ 0.69 & 78.68 & \textbf{92.54}$\pm$ 1.67 & 62.78$\pm$ 0.42 & 80.72$\pm$ 1.40 \\
+ QLoRA Output Summing(IMDB, Yelp) & 85.89 & 91.52 & 91.22$\pm$ 0.19 & 90.00$\pm$ 5.51 & 93.33$\pm$ 2.85 & 80.26 & 89.05$\pm$ 1.72 & 68.06$\pm$ 0.86 & 83.67$\pm$ 1.48 \\
+ QLoRA Output Summing(Yelp, SST-2) & 85.78 & 92.48 & 92.78$\pm$ 1.17 & \cellcolor{gray!25}91.44$\pm$ 2.83 & 93.22$\pm$ 0.38 & 79.07 & 89.05$\pm$ 2.08 & \cellcolor{gray!25}68.94$\pm$ 0.59 & 79.22$\pm$ 1.00 \\
+ QLoRA Output Summing(IMDB, Yelp, SST-2) & \textbf{87.66} & \textbf{94.70} & \textbf{96.33}$\pm$ 1.33 & \textbf{92.67}$\pm$ 1.20 & \cellcolor{gray!25}95.11$\pm$ 1.71 & \cellcolor{gray!25}80.61 & 90.16$\pm$ 0.27 & 66.89$\pm$ 3.47 & \cellcolor{gray!25}84.78$\pm$ 2.06 \\
\dottedline
+ QLoRA Output Averaging(IMDB, SST-2) & 81.35 & 89.74 & 89.22$\pm$ 3.47 & 87.22$\pm$ 1.26 & 92.78$\pm$ 1.17 & 72.95 & 83.02$\pm$ 0.99 & 61.50$\pm$ 1.83 & 74.33$\pm$ 2.62 \\
+ QLoRA Output Averaging(IMDB, Yelp) & 83.73 & 92.30 & 93.00$\pm$ 1.76 & 90.22$\pm$ 3.36 & 93.67$\pm$ 1.76 & 75.16 & 86.03$\pm$ 2.79 & 63.22$\pm$ 2.41 & 76.22$\pm$ 3.15 \\
+ QLoRA Output Averaging(Yelp, SST-2) & 81.03 & 89.00 & 89.67$\pm$ 4.91 & 86.89$\pm$ 2.27 & 90.44$\pm$ 1.35 & 73.06 & 85.08$\pm$ 2.25 & 60.94$\pm$ 2.56 & 73.17$\pm$ 1.42 \\
\dottedline
+ QLoRA Output Averaging(IMDB, Yelp, SST-2) & 80.42 & 89.15 & 88.56$\pm$ 2.12 & 87.11$\pm$ 1.50 & 91.78$\pm$ 0.19 & 71.69 & 85.56$\pm$ 2.62 & 58.39$\pm$ 1.73 & 71.11$\pm$ 3.64 \\
+ QLoRA Averaged Weights(IMDB, SST-2) & 77.64 & 84.07 & 81.00$\pm$ 3.67 & 81.89$\pm$ 3.02 & 89.33$\pm$ 5.13 & 71.21 & 83.97$\pm$ 1.92 & 56.67$\pm$ 2.17 & 73.00$\pm$ 1.04 \\
+ QLoRA Averaged Weights(IMDB, Yelp) & 82.38 & 91.19 & 90.67$\pm$ 2.19 & 88.89$\pm$ 3.66 & 94.00$\pm$ 1.86 & 73.57 & 83.49$\pm$ 3.17 & 61.89$\pm$ 4.03 & 75.33$\pm$ 1.74 \\
+ QLoRA Averaged Weights(Yelp, SST-2) & 77.54 & 84.07 & 81.00$\pm$ 3.67 & 81.89$\pm$ 3.02 & 89.33$\pm$ 5.13 & 71.00 & 83.33$\pm$ 1.65 & 56.67$\pm$ 2.17 & 73.00$\pm$ 1.04 \\
+ QLoRA Averaged Weights(IMDB, Yelp, SST-2) & 81.93 & 89.67 & 91.67$\pm$ 0.58 & 86.67$\pm$ 1.67 & 90.67$\pm$ 1.45 & 74.19 & 85.56$\pm$ 2.25 & 60.72$\pm$ 0.63 & 76.28$\pm$ 1.55 \\
\hline \hline
Mistral 7B & 58.12 & 60.00 & 59.33$\pm$ 0.00 & 55.00$\pm$ 0.00 & 65.67$\pm$ 0.00 & 56.23 & 62.86$\pm$ 0.00 & 52.50$\pm$ 0.00 & 53.33$\pm$ 0.00 \\
+ QLoRA Yelp & \textbf{84.36} & 92.30 & \textbf{94.78}$\pm$ 3.56 & \textbf{90.56}$\pm$ 2.71 & 91.56$\pm$ 2.04 & \textbf{76.42} & 84.60$\pm$ 0.73 & \textbf{61.33}$\pm$ 1.48 & \textbf{83.33}$\pm$ 1.20 \\
+ QLoRA IMDB & 80.51 & 90.63 & 89.67$\pm$ 0.33 & 88.56$\pm$ 4.40 & 93.67$\pm$ 2.96 & 70.38 & 82.70$\pm$ 2.20 & 58.78$\pm$ 1.25 & 69.67$\pm$ 3.92 \\
+ QLoRA SST-2 & 78.77 & 85.67 & 83.11$\pm$ 1.02 & 81.67$\pm$ 1.67 & 92.22$\pm$ 1.07 & 71.87 & 83.65$\pm$ 6.12 & 55.94$\pm$ 0.92 & 76.00$\pm$ 1.64 \\
+ QLoRA Combined Sentiment dataset & 78.81 & 86.30 & 87.11$\pm$ 0.69 & 85.67$\pm$ 5.70 & 86.11$\pm$ 3.75 & 71.32 & 83.17$\pm$ 3.24 & 58.22$\pm$ 0.95 & 72.56$\pm$ 2.43 \\
\hdashline
+ QLoRA Output Summing(IMDB, SST-2) & 80.12 & 87.11 & 85.56$\pm$ 2.55 & 85.44$\pm$ 1.50 & 90.33$\pm$ 3.33 & 73.12 & \cellcolor{gray!25}86.98$\pm$ 3.57 & 57.11$\pm$ 1.78 & 75.28$\pm$ 5.09 \\
+ QLoRA Output Summing(IMDB, Yelp) & 83.02 & 90.44 & 93.33$\pm$ 2.73 & 86.33$\pm$ 3.71 & 91.67$\pm$ 4.10 & \cellcolor{gray!25}75.59 & \cellcolor{gray!25}86.98$\pm$ 2.44 & 58.28$\pm$ 1.00 & \cellcolor{gray!25}81.50$\pm$ 2.33 \\
+ QLoRA Output Summing(Yelp, SST-2) & 81.20 & 88.26 & 89.00$\pm$ 0.58 & 82.33$\pm$ 1.15 & 93.44$\pm$ 1.71 & 74.13 & \textbf{89.68}$\pm$ 1.67 & 59.22$\pm$ 1.51 & 73.50$\pm$ 3.22 \\
+ QLoRA Output Summing(IMDB, Yelp, SST-2) & \cellcolor{gray!25}83.45 & \textbf{92.81} & \cellcolor{gray!25}93.78$\pm$ 1.95 & 89.44$\pm$ 1.26 & \textbf{95.22}$\pm$ 2.22 & 74.09 & 86.67$\pm$ 1.26 & 58.89$\pm$ 0.84 & 76.72$\pm$ 2.18 \\
\dottedline
+ QLoRA Output Averaging(IMDB, SST-2) & 79.41 & 88.89 & 88.56$\pm$ 1.07 & 85.33$\pm$ 1.53 & 92.78$\pm$ 1.02 & 69.93 & 82.06$\pm$ 0.27 & 58.39$\pm$ 0.42 & 69.33$\pm$ 2.62 \\
+ QLoRA Output Averaging(IMDB, Yelp) & 82.49 & \cellcolor{gray!25}92.63 & 93.56$\pm$ 1.07 & \cellcolor{gray!25}90.11$\pm$ 1.02 & 94.22$\pm$ 0.77 & 72.35 & 84.60$\pm$ 2.75 & 56.78$\pm$ 0.35 & 75.67$\pm$ 0.83 \\
+ QLoRA Output Averaging(Yelp, SST-2) & 81.35 & 90.74 & 92.44$\pm$ 1.54 & 86.33$\pm$ 3.84 & 93.44$\pm$ 0.84 & 71.96 & 85.87$\pm$ 0.99 & \cellcolor{gray!25}60.67$\pm$ 0.73 & 69.33$\pm$ 2.25 \\
+ QLoRA Output Averaging(IMDB, Yelp, SST-2) & 81.68 & 92.37 & 91.67$\pm$ 0.67 & \textbf{90.56}$\pm$ 2.71 & \cellcolor{gray!25}94.89$\pm$ 0.51 & 70.99 & 84.76$\pm$ 4.15 & 57.67$\pm$ 2.18 & 70.56$\pm$ 0.84 \\
\dottedline
+ QLoRA Averaged Weights(IMDB, SST-2) & 78.92 & 85.85 & 82.89$\pm$ 1.17 & 82.22$\pm$ 2.01 & 92.44$\pm$ 1.50 & 71.98 & 84.60$\pm$ 2.91 & 56.28$\pm$ 0.51 & 75.06$\pm$ 3.23 \\
+ QLoRA Averaged Weights(IMDB, Yelp) & 81.03 & 91.33 & 90.22$\pm$ 1.71 & 89.89$\pm$ 4.86 & 93.89$\pm$ 2.78 & 70.72 & 83.65$\pm$ 2.25 & 59.56$\pm$ 1.84 & 68.94$\pm$ 2.51 \\
+ QLoRA Averaged Weights(Yelp, SST-2) & 79.23 & 85.85 & 82.89$\pm$ 1.17 & 82.22$\pm$ 2.01 & 92.44$\pm$ 1.50 & 72.61 & 86.51$\pm$ 4.50 & 56.28$\pm$ 0.51 & 75.06$\pm$ 3.23 \\
+ QLoRA Averaged Weights(IMDB, Yelp, SST-2) & 79.97 & 88.26 & 88.11$\pm$ 2.27 & 84.11$\pm$ 0.69 & 92.56$\pm$ 0.77 & 71.67 & 86.35$\pm$ 3.10 & 58.00$\pm$ 0.44 & 70.67$\pm$ 2.84 \\
\hline
    \end{tabular}
    \caption{\textbf{Sentiment Control} Control Effectiveness for the model + QLoRA module combinations explained in Section~\ref{sec:study_overview}. Here, e.g.\ Output Summing(data1, data2) refers to the output summation module composition technique. All values are averages over 3 runs and standard deviation is reported for single datasets results.
    Bold (shaded) =  (second) highest score in column/section; underline = train/test on same dataset.
    }
    \label{tab:sent_res_std}
\end{table*}

\begin{table*}[h]
    \centering
    \small
    \setlength\tabcolsep{2pt} % default value: 6pt
    \renewcommand{\arraystretch}{1.10}
    \begin{tabular}{|l|ccc|c|}
    \hline
         \multirow{2}{*}{\textbf{CTG Technique}} & \multicolumn{3}{c|}{\textbf{Distinct-n}$\uparrow$} & \multirow{2}{*}{\textbf{SLOR}$\uparrow$} \\ \cline{2-4}
         & \textbf{dist-1} & \textbf{dist-2} & \textbf{dist-3} &  \\
\hline
Llama 3 8B & 0.07$\pm$ 0.02 & 0.17$\pm$ 0.05 & 0.22$\pm$ 0.05 & 8.45$\pm$ 0.74 \\
+ QLoRA AG News & 0.25$\pm$ 0.09 & 0.53$\pm$ 0.15 & 0.62$\pm$ 0.16 & \cellcolor{gray!25}9.39$\pm$ 0.13 \\
+ QLoRA DBPedia & 0.32$\pm$ 0.09 & 0.60$\pm$ 0.13 & 0.68$\pm$ 0.13 & 8.96$\pm$ 0.37 \\
+ QLoRA Combined Topic dataset & 0.30$\pm$ 0.10 & 0.60$\pm$ 0.15 & 0.70$\pm$ 0.15 & 8.92$\pm$ 0.30 \\
\hdashline
+ QLoRA Output Summing(AG News, DBPedia) & \textbf{0.35}$\pm$ 0.08 & \textbf{0.70}$\pm$ 0.11 & \textbf{0.80}$\pm$ 0.09 & 9.33$\pm$ 0.36 \\
+ QLoRA Output Averaging(AG News, DBPedia) & 0.27$\pm$ 0.10 & 0.55$\pm$ 0.17 & 0.64$\pm$ 0.18 & \textbf{9.57}$\pm$ 0.17 \\
+ QLoRA Averaged Weights(AG News, DBPedia) & \cellcolor{gray!25}0.33$\pm$ 0.08 & \cellcolor{gray!25}0.62$\pm$ 0.11 & \cellcolor{gray!25}0.71$\pm$ 0.12 & 8.98$\pm$ 0.41 \\
\hline \hline
Llama 3.1 8B & 0.05$\pm$ 0.02 & 0.12$\pm$ 0.04 & 0.17$\pm$ 0.03 & \textbf{9.45}$\pm$ 0.76 \\
+ QLoRA AG News & 0.26$\pm$ 0.09 & 0.55$\pm$ 0.15 & 0.65$\pm$ 0.15 & \cellcolor{gray!25}9.40$\pm$ 0.11 \\
+ QLoRA DBPedia & 0.31$\pm$ 0.08 & 0.59$\pm$ 0.11 & 0.68$\pm$ 0.11 & 8.74$\pm$ 0.45 \\
+ QLoRA Combined Topic dataset & 0.31$\pm$ 0.10 & \cellcolor{gray!25}0.62$\pm$ 0.16 & \cellcolor{gray!25}0.73$\pm$ 0.16 & 9.29$\pm$ 0.20 \\
\hdashline
+ QLoRA Output Summing(AG News, DBPedia) & \textbf{0.35}$\pm$ 0.07 & \textbf{0.68}$\pm$ 0.10 & \textbf{0.78}$\pm$ 0.09 & 9.12$\pm$ 0.39 \\
+ QLoRA Output Averaging(AG News, DBPedia) & 0.30$\pm$ 0.09 & 0.59$\pm$ 0.14 & 0.68$\pm$ 0.14 & 8.98$\pm$ 0.23 \\
+ QLoRA Averaged Weights(AG News, DBPedia) & \cellcolor{gray!25}0.32$\pm$ 0.08 & 0.61$\pm$ 0.11 & 0.71$\pm$ 0.10 & 8.74$\pm$ 0.37 \\
\hline \hline
Mistral 7B & 0.07$\pm$ 0.05 & 0.13$\pm$ 0.07 & 0.16$\pm$ 0.06 & 7.16$\pm$ 2.31 \\
+ QLoRA AG News & 0.18$\pm$ 0.10 & 0.37$\pm$ 0.17 & 0.44$\pm$ 0.18 & \cellcolor{gray!25}9.48$\pm$ 0.28 \\
+ QLoRA DBPedia & \textbf{0.26}$\pm$ 0.06 & \cellcolor{gray!25}0.48$\pm$ 0.08 & \cellcolor{gray!25}0.56$\pm$ 0.08 & 8.81$\pm$ 0.54 \\
+ QLoRA Combined Topic dataset & \cellcolor{gray!25}0.21$\pm$ 0.09 & 0.43$\pm$ 0.14 & 0.52$\pm$ 0.15 & 9.09$\pm$ 0.31 \\
\hdashline
+ QLoRA Output Summing(AG News, DBPedia) & \textbf{0.26}$\pm$ 0.09 & \textbf{0.50}$\pm$ 0.15 & \textbf{0.59}$\pm$ 0.15 & 9.26$\pm$ 0.17 \\
+ QLoRA Output Averaging(AG News, DBPedia) & 0.18$\pm$ 0.10 & 0.36$\pm$ 0.16 & 0.42$\pm$ 0.18 & \textbf{9.60}$\pm$ 0.13 \\
+ QLoRA Averaged Weights(AG News, DBPedia) & \textbf{0.26}$\pm$ 0.06 & \cellcolor{gray!25}0.48$\pm$ 0.08 & \cellcolor{gray!25}0.56$\pm$ 0.08 & 8.87$\pm$ 0.62 \\
\hline
    \end{tabular}
    \caption{Diversity, Fluency for \textbf{Topic Control}, training on \textit{single} and \textit{combined} datasets, and composition of modules trained on single datasets, e.g.\ Output Summing(data1, data2). All values are averages over 3 runs and standard deviation is reported. Bold (shaded) =  (second) highest score in column and section.}
    \label{tab:topic_res_std_dist}
\end{table*}

\begin{table*}[h]
    \centering
    \scriptsize
    \setlength\tabcolsep{2pt} % default value: 6pt
    \renewcommand{\arraystretch}{1.10}
    \begin{tabular}{|l|c|ccc|cccc|}
    \hline
         \multirow{3}{*}{\textbf{CTG Technique}} & \multicolumn{8}{|c|}{\textbf{Control Effectiveness}$\uparrow$} \\ \cline{2-9}
         & \multirow{2}{*}{\textbf{Avg All}} & \multirow{2}{*}{\textbf{Avg}} & \multirow{2}{*}{\textbf{AG News}} & \multirow{2}{*}{\textbf{DBPedia}} & \multicolumn{4}{c|}{\textbf{Out-Of-Domain}} \\ \cline{6-9}
         &  &  &  &  & \textbf{Avg} & \textbf{PPLM T} & \textbf{STS T} & \textbf{STS proc T} \\
\hline
Llama 3 8B & 45.92 & 58.52 & 64.61$\pm$ 2.43 & 52.42$\pm$ 2.32 & 37.53 & 48.97$\pm$ 3.49 & 27.97$\pm$ 1.35 & 35.64$\pm$ 1.25 \\
+ QLoRA AG News & \textbf{68.86} & \textbf{85.13} & \textbf{90.72}$\pm$ 0.42 & \textbf{79.54}$\pm$ 0.99 & \textbf{58.02} & \textbf{71.11}$\pm$ 4.33 & \textbf{42.03}$\pm$ 0.39 & \cellcolor{gray!25}60.92$\pm$ 2.21 \\
+ QLoRA DBPedia & 52.97 & 69.94 & 71.67$\pm$ 1.61 & 68.21$\pm$ 2.22 & 41.66 & 55.40$\pm$ 1.45 & 28.58$\pm$ 1.63 & 41.00$\pm$ 1.80 \\
+ QLoRA Combined Topic dataset & 63.53 & 74.42 & 81.61$\pm$ 1.11 & 67.24$\pm$ 1.45 & \cellcolor{gray!25}56.27 & \cellcolor{gray!25}68.73$\pm$ 4.73 & \cellcolor{gray!25}37.50$\pm$ 2.11 & \textbf{62.58}$\pm$ 0.17 \\
\hdashline
+ QLoRA Output Summing(AG News, DBPedia) & \cellcolor{gray!25}64.58 & \cellcolor{gray!25}82.57 & \cellcolor{gray!25}88.78$\pm$ 0.35 & \cellcolor{gray!25}76.35$\pm$ 0.43 & 52.59 & \textbf{71.11}$\pm$ 1.31 & 32.81$\pm$ 0.43 & 53.86$\pm$ 1.24 \\
+ QLoRA Output Averaging(AG News, DBPedia) & 61.75 & 78.33 & 83.78$\pm$ 2.17 & 72.88$\pm$ 0.94 & 50.70 & 66.98$\pm$ 1.59 & 33.81$\pm$ 1.95 & 51.31$\pm$ 1.85 \\
+ QLoRA Averaged Weights(AG News, DBPedia) & 55.55 & 70.61 & 72.11$\pm$ 1.92 & 69.12$\pm$ 3.08 & 45.50 & 66.59$\pm$ 0.90 & 28.06$\pm$ 1.85 & 41.86$\pm$ 1.64 \\
\hline \hline
Llama 3.1 8B & 45.93 & 58.61 & 66.11$\pm$ 1.21 & 51.11$\pm$ 2.52 & 37.47 & 46.35$\pm$ 2.21 & 29.89$\pm$ 0.49 & 36.17$\pm$ 0.36 \\
+ QLoRA AG News & \textbf{68.53} & \textbf{85.52} & \textbf{91.33}$\pm$ 1.50 & \textbf{79.72}$\pm$ 0.36 & \textbf{57.21} & \textbf{73.10}$\pm$ 2.58 & \cellcolor{gray!25}36.28$\pm$ 0.34 & \textbf{62.25}$\pm$ 0.66 \\
+ QLoRA DBPedia & 52.93 & 69.86 & 70.94$\pm$ 0.92 & 68.77$\pm$ 3.27 & 41.64 & 54.92$\pm$ 2.55 & 28.11$\pm$ 1.10 & 41.89$\pm$ 1.97 \\
+ QLoRA Combined Topic dataset & \cellcolor{gray!25}65.27 & 80.65 & \cellcolor{gray!25}89.06$\pm$ 0.69 & 72.25$\pm$ 2.23 & \cellcolor{gray!25}55.01 & 66.51$\pm$ 1.37 & \textbf{39.97}$\pm$ 6.45 & \cellcolor{gray!25}58.56$\pm$ 5.01 \\
\hdashline
+ QLoRA Output Summing(AG News, DBPedia) & 64.15 & \cellcolor{gray!25}82.91 & 88.39$\pm$ 1.36 & \cellcolor{gray!25}77.44$\pm$ 0.89 & 51.64 & \cellcolor{gray!25}69.05$\pm$ 2.42 & 29.39$\pm$ 2.19 & 56.47$\pm$ 0.97 \\
+ QLoRA Output Averaging(AG News, DBPedia) & 59.21 & 76.42 & 83.44$\pm$ 2.07 & 69.40$\pm$ 1.04 & 47.74 & 61.67$\pm$ 1.09 & 29.81$\pm$ 2.16 & 51.75$\pm$ 1.53 \\
+ QLoRA Averaged Weights(AG News, DBPedia) & 53.64 & 69.52 & 70.94$\pm$ 1.00 & 68.09$\pm$ 2.57 & 43.06 & 60.71$\pm$ 0.95 & 27.61$\pm$ 0.64 & 40.86$\pm$ 1.47 \\
\hline \hline
Mistral 7B & 44.87 & 55.96 & 62.00$\pm$ 0.00 & 49.91$\pm$ 0.00 & 37.48 & 49.76$\pm$ 0.00 & 29.92$\pm$ 0.00 & 32.75$\pm$ 0.00 \\
+ QLoRA AG News & \textbf{69.69} & \textbf{88.52} & \textbf{93.17}$\pm$ 0.88 & \textbf{83.87}$\pm$ 1.38 & \textbf{57.14} & \textbf{73.97}$\pm$ 2.50 & \textbf{39.69}$\pm$ 2.71 & \textbf{57.75}$\pm$ 0.80 \\
+ QLoRA DBPedia & 53.08 & 67.05 & 67.50$\pm$ 1.04 & 66.61$\pm$ 0.94 & 43.76 & 54.13$\pm$ 0.50 & 32.89$\pm$ 0.39 & 44.28$\pm$ 1.84 \\
+ QLoRA Combined Topic dataset & 65.02 & 81.74 & \cellcolor{gray!25}89.00$\pm$ 2.35 & 74.47$\pm$ 1.11 & \cellcolor{gray!25}53.87 & 67.54$\pm$ 3.86 & \cellcolor{gray!25}39.22$\pm$ 2.88 & \cellcolor{gray!25}54.86$\pm$ 4.65 \\
\hdashline
+ QLoRA Output Summing(AG News, DBPedia) & \cellcolor{gray!25}66.02 & \cellcolor{gray!25}85.27 & 88.89$\pm$ 0.54 & \cellcolor{gray!25}81.65$\pm$ 1.31 & 53.19 & \cellcolor{gray!25}69.76$\pm$ 2.18 & 36.36$\pm$ 0.69 & 53.44$\pm$ 2.35 \\
+ QLoRA Output Averaging(AG News, DBPedia) & 58.39 & 76.01 & 82.00$\pm$ 3.53 & 70.03$\pm$ 0.94 & 46.64 & 60.95$\pm$ 1.09 & 33.89$\pm$ 0.94 & 45.08$\pm$ 0.55 \\
+ QLoRA Averaged Weights(AG News, DBPedia) & 54.65 & 66.53 & 66.06$\pm$ 1.17 & 67.01$\pm$ 0.51 & 46.72 & 61.75$\pm$ 1.76 & 33.47$\pm$ 0.98 & 44.94$\pm$ 2.96 \\
\hline
    \end{tabular}
    \caption{Diversity, Fluency, Control Effectiveness for \textbf{Topic Control}, training on \textit{single} and \textit{combined} datasets, and composition of modules trained on single datasets, e.g.\ Output Summing(data1, data2). All values are averages over 3 runs and standard deviation is reported for single dataset results. Bold (shaded) =  (second) highest score in column and section; underline = train and test set from same dataset.}
    \label{tab:topic_res_std_ce}
\end{table*}

\begin{table*}[h]
    \centering
    \small
    \setlength\tabcolsep{1pt} % default value: 6pt
    \renewcommand{\arraystretch}{1.10}
    \begin{tabular}{|l|ccc|c|}
\hline
\multirow{2}{*}{\textbf{CTG Technique}} & \multicolumn{3}{c|}{\textbf{Distinct-n}$\uparrow$} & \multirow{2}{*}{\textbf{SLOR}$\uparrow$} \\ \cline{2-4}
& \textbf{dist1} & \textbf{dist2} & \textbf{dist3} &  \\
\hline
Llama 3 8B & 0.03 $\pm$ 0.00 & 0.10 $\pm$ 0.01 & 0.14 $\pm$ 0.01 & 8.36 $\pm$ 0.11 \\
+ QLoRA Combined S & 0.03 $\pm$ 0.00 & 0.09 $\pm$ 0.01 & 0.14 $\pm$ 0.01 & \textbf{10.81 $\pm$ 0.12} \\
+ QLoRA Combined T & \textbf{0.24 $\pm$ 0.02} & \textbf{0.50 $\pm$ 0.04} & \textbf{0.59 $\pm$ 0.05} & 8.52 $\pm$ 0.20 \\
\hdashline
+ QLoRA Sum(Ind mod) & 0.18 $\pm$ 0.01 & 0.42 $\pm$ 0.03 & 0.51 $\pm$ 0.04 & 9.11 $\pm$ 0.07 \\
+ QLoRA Sum(S, T) & 0.09 $\pm$ 0.01 & 0.23 $\pm$ 0.02 & 0.31 $\pm$ 0.03 & \cellcolor{gray!25}9.81 $\pm$ 0.24 \\
\dottedline
+ QLoRA Average(Ind mod) & 0.08 $\pm$ 0.01 & 0.20 $\pm$ 0.01 & 0.26 $\pm$ 0.01 & 9.55 $\pm$ 0.05 \\
+ QLoRA Average(S, T) & 0.10 $\pm$ 0.03 & 0.24 $\pm$ 0.07 & 0.31 $\pm$ 0.08 & 9.26 $\pm$ 0.21 \\
\dottedline
+ QLoRA Weights Average(Ind mod) & 0.09 $\pm$ 0.01 & 0.23 $\pm$ 0.03 & 0.30 $\pm$ 0.03 & 9.45 $\pm$ 0.10 \\
+ QLoRA Weights Average(S, T) & \cellcolor{gray!25}0.21 $\pm$ 0.04 & \cellcolor{gray!25}0.46 $\pm$ 0.07 & \cellcolor{gray!25}0.55 $\pm$ 0.07 & 8.56 $\pm$ 0.24 \\
\hline \hline
Llama 3.1 8B & 0.05 $\pm$ 0.00 & 0.12 $\pm$ 0.01 & 0.18 $\pm$ 0.01 & \cellcolor{gray!25}9.84 $\pm$ 0.09 \\
+ QLoRA Combined S & 0.03 $\pm$ 0.01 & 0.11 $\pm$ 0.02 & 0.17 $\pm$ 0.03 & \textbf{10.73 $\pm$ 0.17} \\
+ QLoRA Combined T & \textbf{0.28 $\pm$ 0.05} & \textbf{0.58 $\pm$ 0.07} & \textbf{0.68 $\pm$ 0.08} & 8.58 $\pm$ 0.50 \\
\hdashline
+ QLoRA Sum(Ind mod) & 0.13 $\pm$ 0.02 & 0.29 $\pm$ 0.05 & 0.35 $\pm$ 0.06 & 8.63 $\pm$ 0.16 \\
+ QLoRA Sum(S, T) & 0.10 $\pm$ 0.07 & 0.27 $\pm$ 0.12 & 0.36 $\pm$ 0.13 & 9.80 $\pm$ 0.49 \\
\dottedline
+ QLoRA Average(Ind mod) & 0.08 $\pm$ 0.01 & 0.22 $\pm$ 0.02 & 0.29 $\pm$ 0.02 & 9.36 $\pm$ 0.08 \\
+ QLoRA Average(S, T) & 0.11 $\pm$ 0.07 & 0.29 $\pm$ 0.13 & 0.37 $\pm$ 0.15 & 9.67 $\pm$ 0.57 \\
\dottedline
+ QLoRA Weights Average(Ind mod) & 0.10 $\pm$ 0.01 & 0.25 $\pm$ 0.03 & 0.33 $\pm$ 0.03 & 9.14 $\pm$ 0.04 \\
+ QLoRA Weights Average(S, T) & \cellcolor{gray!25}0.24 $\pm$ 0.07 & \cellcolor{gray!25}0.52 $\pm$ 0.13 & \cellcolor{gray!25}0.62 $\pm$ 0.15 & 8.74 $\pm$ 0.63 \\
\hline \hline
Mistral 7B & 0.03 $\pm$ 0.00 & 0.07 $\pm$ 0.00 & 0.10 $\pm$ 0.00 & 9.77 $\pm$ 0.00 \\
+ QLoRA Combined S & 0.01 $\pm$ 0.00 & 0.03 $\pm$ 0.00 & 0.05 $\pm$ 0.00 & \textbf{11.60 $\pm$ 0.11} \\
+ QLoRA Combined T & \textbf{0.12 $\pm$ 0.03} & \textbf{0.27 $\pm$ 0.08} & \textbf{0.36 $\pm$ 0.11} & 8.92 $\pm$ 0.31 \\
\hdashline
+ QLoRA Sum(Ind mod) & 0.09 $\pm$ 0.06 & 0.20 $\pm$ 0.10 & 0.24 $\pm$ 0.11 & 8.94 $\pm$ 0.54 \\
+ QLoRA Sum(S, T) & 0.09 $\pm$ 0.07 & 0.22 $\pm$ 0.16 & 0.30 $\pm$ 0.21 & 9.50 $\pm$ 1.23 \\
\dottedline
+ QLoRA Average(Ind mod) & 0.02 $\pm$ 0.00 & 0.05 $\pm$ 0.00 & 0.08 $\pm$ 0.00 & \cellcolor{gray!25}10.75 $\pm$ 0.20 \\
+ QLoRA Average(S, T) & 0.08 $\pm$ 0.07 & 0.18 $\pm$ 0.14 & 0.23 $\pm$ 0.16 & 9.63 $\pm$ 1.01 \\
\dottedline
+ QLoRA Weights Average(Ind mod) & 0.03 $\pm$ 0.00 & 0.07 $\pm$ 0.01 & 0.10 $\pm$ 0.01 & 10.18 $\pm$ 0.35 \\
+ QLoRA Weights Average(S, T) & \cellcolor{gray!25}0.10 $\pm$ 0.03 & \cellcolor{gray!25}0.24 $\pm$ 0.08 & \cellcolor{gray!25}0.32 $\pm$ 0.11 & 9.04 $\pm$ 0.48 \\
\hline
\end{tabular}
\caption{Diversity, Fluency \textbf{Multi-attribute Control} alongside single-attribute control results for comparison. All values are averages over 3 runs and standard deviation is reported. S=the Combined Sentiment dataset, T=the Combined Topic dataset, Ind mod=composition is on all 5 individually trained modules. Bold (shaded) =  (second) highest score in column and section; underline = train and test set from same dataset.
} 
\label{tab:multi_res_std_dist}
\end{table*}

\begin{table*}[h]
    \centering
    \scriptsize
    \setlength\tabcolsep{1pt} % default value: 6pt
    \renewcommand{\arraystretch}{1.10}
    \begin{tabular}{|l|c|ccc|ccc|ccc|}
\hline
\multirow{4}{*}{\textbf{CTG Technique}} & \multicolumn{10}{c|}{\textbf{Control Effectiveness} $\uparrow$} \\ \cline{2-11}
& \multicolumn{10}{c|}{\textbf{Out-Of-Domain}} \\ \cline{2-11}
& \multicolumn{4}{c|}{\textbf{Multiple}} & \multicolumn{3}{c|}{\textbf{Sentiment}} & \multicolumn{3}{c|}{\textbf{Topic}} \\
\cline{2-11}
& \textbf{Avg} & \textbf{PPLM M} & \textbf{STS M} & \textbf{STS p M} &\textbf{PPLM S} & \textbf{STS S} & \textbf{STS p S} & \textbf{PPLM T} & \textbf{STS T} & \textbf{STS p T} \\
\hline
Llama 3 8B & 19.22 & 29.29 $\pm$ 0.02 & 14.54 $\pm$ 0.00 & 20.38 $\pm$ 0.01 & 0.65 $\pm$ 0.04 & 0.53 $\pm$ 0.01 & 0.54 $\pm$ 0.02 & 0.49 $\pm$ 0.03 & 0.28 $\pm$ 0.01 & 0.36 $\pm$ 0.01 \\
+ QLoRA Combined S & \textbf{25.66} & \cellcolor{gray!25}37.14 $\pm$ 0.03 & 16.50 $\pm$ 0.01 & \textbf{30.79 $\pm$ 0.02} & \cellcolor{gray!25}0.87 $\pm$ 0.02 & \textbf{0.64 $\pm$ 0.02} & \textbf{0.77 $\pm$ 0.03} & 0.52 $\pm$ 0.02 & 0.28 $\pm$ 0.02 & 0.38 $\pm$ 0.01 \\
+ QLoRA Combined T & 22.64 & 36.55 $\pm$ 0.03 & 14.96 $\pm$ 0.01 & 25.46 $\pm$ 0.00 & 0.56 $\pm$ 0.02 & 0.51 $\pm$ 0.00 & 0.55 $\pm$ 0.02 & \textbf{0.69 $\pm$ 0.05} & \textbf{0.38 $\pm$ 0.02} & \textbf{0.63 $\pm$ 0.00} \\
\hdashline
+ QLoRA Sum(Ind mod) & \cellcolor{gray!25}24.08 & 33.69 $\pm$ 0.02 & \cellcolor{gray!25}16.79 $\pm$ 0.01 & \cellcolor{gray!25}28.00 $\pm$ 0.01 & 0.80 $\pm$ 0.05 & 0.59 $\pm$ 0.05 & 0.69 $\pm$ 0.01 & 0.44 $\pm$ 0.06 & 0.25 $\pm$ 0.01 & 0.38 $\pm$ 0.05 \\
+ QLoRA Sum(S, T) & 23.85 & 35.60 $\pm$ 0.03 & \textbf{17.75 $\pm$ 0.01} & 25.83 $\pm$ 0.02 & \textbf{0.88 $\pm$ 0.04} & \cellcolor{gray!25}0.60 $\pm$ 0.01 & \cellcolor{gray!25}0.73 $\pm$ 0.03 & 0.62 $\pm$ 0.07 & 0.29 $\pm$ 0.02 & 0.50 $\pm$ 0.02 \\
\dottedline
+ QLoRA Average(Ind mod) & 23.30 & \textbf{37.50 $\pm$ 0.03} & 15.46 $\pm$ 0.01 & 26.17 $\pm$ 0.00 & 0.78 $\pm$ 0.04 & 0.57 $\pm$ 0.01 & 0.68 $\pm$ 0.05 & 0.58 $\pm$ 0.02 & 0.27 $\pm$ 0.01 & 0.43 $\pm$ 0.01 \\
+ QLoRA Average(S, T) & 22.32 & 36.55 $\pm$ 0.03 & 14.50 $\pm$ 0.01 & 25.17 $\pm$ 0.02 & 0.76 $\pm$ 0.07 & 0.55 $\pm$ 0.01 & 0.64 $\pm$ 0.03 & 0.58 $\pm$ 0.05 & 0.29 $\pm$ 0.02 & 0.42 $\pm$ 0.01 \\
\dottedline
+ QLoRA Weights Average(Ind mod) & 23.58 & 35.71 $\pm$ 0.04 & 15.79 $\pm$ 0.01 & 27.12 $\pm$ 0.01 & 0.74 $\pm$ 0.03 & 0.54 $\pm$ 0.03 & 0.63 $\pm$ 0.03 & 0.65 $\pm$ 0.02 & 0.32 $\pm$ 0.02 & 0.46 $\pm$ 0.02 \\
+ QLoRA Weights Average(S, T) & 22.85 & 36.55 $\pm$ 0.04 & 15.62 $\pm$ 0.01 & 25.29 $\pm$ 0.00 & 0.75 $\pm$ 0.01 & 0.51 $\pm$ 0.00 & 0.55 $\pm$ 0.02 & \cellcolor{gray!25}0.68 $\pm$ 0.03 & \cellcolor{gray!25}0.37 $\pm$ 0.02 & \cellcolor{gray!25}0.60 $\pm$ 0.02 \\
\hline \hline
Llama 3.1 8B & 19.95 & 28.57 $\pm$ 0.02 & 15.00 $\pm$ 0.01 & 21.88 $\pm$ 0.01 & 0.64 $\pm$ 0.03 & 0.52 $\pm$ 0.02 & 0.53 $\pm$ 0.04 & 0.46 $\pm$ 0.02 & 0.30 $\pm$ 0.00 & 0.36 $\pm$ 0.00 \\
+ QLoRA Combined S & \cellcolor{gray!25}26.44 & 37.50 $\pm$ 0.04 & \cellcolor{gray!25}17.54 $\pm$ 0.01 & \textbf{31.46 $\pm$ 0.01} & \textbf{0.90 $\pm$ 0.05} & \cellcolor{gray!25}0.62 $\pm$ 0.01 & \textbf{0.82 $\pm$ 0.03} & 0.49 $\pm$ 0.07 & 0.31 $\pm$ 0.01 & 0.37 $\pm$ 0.01 \\
+ QLoRA Combined T & 23.95 & 35.12 $\pm$ 0.05 & \cellcolor{gray!25}17.54 $\pm$ 0.07 & 26.46 $\pm$ 0.07 & 0.52 $\pm$ 0.05 & 0.51 $\pm$ 0.02 & 0.52 $\pm$ 0.02 & \cellcolor{gray!25}0.67 $\pm$ 0.01 & \textbf{0.40 $\pm$ 0.06} & \cellcolor{gray!25}0.59 $\pm$ 0.05 \\
\hdashline
+ QLoRA Sum(Ind mod) & 20.78 & 30.24 $\pm$ 0.03 & 16.29 $\pm$ 0.01 & 21.96 $\pm$ 0.02 & \cellcolor{gray!25}0.85 $\pm$ 0.03 & 0.61 $\pm$ 0.02 & \cellcolor{gray!25}0.73 $\pm$ 0.01 & 0.35 $\pm$ 0.04 & 0.24 $\pm$ 0.01 & 0.29 $\pm$ 0.02 \\
+ QLoRA Sum(S, T) & \textbf{26.76} & 34.40 $\pm$ 0.03 & \textbf{19.58 $\pm$ 0.03} & \cellcolor{gray!25}31.25 $\pm$ 0.03 & 0.76 $\pm$ 0.20 & \textbf{0.63 $\pm$ 0.05} & 0.72 $\pm$ 0.13 & 0.58 $\pm$ 0.14 & \cellcolor{gray!25}0.35 $\pm$ 0.11 & 0.50 $\pm$ 0.15 \\
\dottedline
+ QLoRA Average(Ind mod) & 23.17 & 37.38 $\pm$ 0.03 & 15.21 $\pm$ 0.00 & 26.17 $\pm$ 0.00 & 0.70 $\pm$ 0.01 & 0.55 $\pm$ 0.02 & 0.65 $\pm$ 0.02 & 0.46 $\pm$ 0.01 & 0.27 $\pm$ 0.01 & 0.39 $\pm$ 0.00 \\
+ QLoRA Average(S, T) & 25.89 & \textbf{40.83 $\pm$ 0.07} & 16.50 $\pm$ 0.02 & 30.04 $\pm$ 0.01 & 0.74 $\pm$ 0.12 & 0.60 $\pm$ 0.05 & 0.64 $\pm$ 0.09 & 0.59 $\pm$ 0.04 & 0.30 $\pm$ 0.04 & 0.43 $\pm$ 0.09 \\
\dottedline
+ QLoRA Weights Average(Ind mod) & 23.26 & \cellcolor{gray!25}40.36 $\pm$ 0.04 & 15.46 $\pm$ 0.01 & 25.08 $\pm$ 0.01 & 0.72 $\pm$ 0.04 & 0.54 $\pm$ 0.03 & 0.64 $\pm$ 0.01 & 0.50 $\pm$ 0.02 & 0.28 $\pm$ 0.01 & 0.40 $\pm$ 0.02 \\
+ QLoRA Weights Average(S, T) & 24.43 & 37.98 $\pm$ 0.06 & 17.29 $\pm$ 0.07 & 26.83 $\pm$ 0.05 & 0.73 $\pm$ 0.14 & 0.51 $\pm$ 0.02 & 0.52 $\pm$ 0.02 & \textbf{0.70 $\pm$ 0.03} & \textbf{0.40 $\pm$ 0.07} & \textbf{0.61 $\pm$ 0.05} \\
\hline \hline
Mistral 7B & 20.43 & 26.07 $\pm$ 0.00 & 16.12 $\pm$ 0.00 & 22.75 $\pm$ 0.00 & 0.63 $\pm$ 0.00 & 0.53 $\pm$ 0.00 & 0.53 $\pm$ 0.00 & 0.50 $\pm$ 0.00 & 0.30 $\pm$ 0.00 & 0.33 $\pm$ 0.00 \\
+ QLoRA Combined S & 21.47 & 30.60 $\pm$ 0.01 & 14.75 $\pm$ 0.01 & 25.00 $\pm$ 0.01 & \cellcolor{gray!25}0.83 $\pm$ 0.03 & \textbf{0.58 $\pm$ 0.01} & \cellcolor{gray!25}0.73 $\pm$ 0.02 & 0.47 $\pm$ 0.04 & 0.27 $\pm$ 0.01 & 0.38 $\pm$ 0.02 \\
+ QLoRA Combined T & \cellcolor{gray!25}26.06 & 35.12 $\pm$ 0.04 & \cellcolor{gray!25}20.33 $\pm$ 0.03 & \cellcolor{gray!25}28.62 $\pm$ 0.03 & 0.50 $\pm$ 0.03 & 0.52 $\pm$ 0.01 & 0.52 $\pm$ 0.01 & \cellcolor{gray!25}0.68 $\pm$ 0.04 & \cellcolor{gray!25}0.39 $\pm$ 0.03 & 0.55 $\pm$ 0.05 \\
\hdashline
+ QLoRA Sum(Ind mod) & 22.77 & \cellcolor{gray!25}36.19 $\pm$ 0.03 & 16.25 $\pm$ 0.01 & 24.58 $\pm$ 0.02 & \textbf{0.88 $\pm$ 0.02} & \cellcolor{gray!25}0.57 $\pm$ 0.03 & \textbf{0.76 $\pm$ 0.02} & 0.57 $\pm$ 0.04 & 0.31 $\pm$ 0.01 & 0.42 $\pm$ 0.02 \\
+ QLoRA Sum(S, T) & 25.53 & 34.05 $\pm$ 0.03 & 19.58 $\pm$ 0.04 & 28.50 $\pm$ 0.03 & 0.60 $\pm$ 0.15 & 0.51 $\pm$ 0.05 & 0.58 $\pm$ 0.12 & 0.63 $\pm$ 0.01 & \textbf{0.40 $\pm$ 0.01} & \textbf{0.61 $\pm$ 0.01} \\
\dottedline
+ QLoRA Average(Ind mod) & 23.01 & 35.95 $\pm$ 0.01 & 14.08 $\pm$ 0.01 & 27.42 $\pm$ 0.02 & 0.76 $\pm$ 0.02 & 0.55 $\pm$ 0.02 & 0.67 $\pm$ 0.03 & 0.48 $\pm$ 0.03 & 0.28 $\pm$ 0.01 & 0.38 $\pm$ 0.01 \\
+ QLoRA Average(S, T) & 22.73 & 34.88 $\pm$ 0.07 & 15.54 $\pm$ 0.02 & 25.67 $\pm$ 0.02 & 0.63 $\pm$ 0.11 & 0.53 $\pm$ 0.05 & 0.57 $\pm$ 0.06 & 0.61 $\pm$ 0.03 & 0.30 $\pm$ 0.00 & 0.44 $\pm$ 0.04 \\
\dottedline
+ QLoRA Weights Average(Ind mod) & 21.70 & \textbf{36.31 $\pm$ 0.01} & 14.42 $\pm$ 0.00 & 23.88 $\pm$ 0.02 & 0.77 $\pm$ 0.01 & 0.54 $\pm$ 0.00 & 0.62 $\pm$ 0.00 & 0.53 $\pm$ 0.01 & 0.28 $\pm$ 0.00 & 0.39 $\pm$ 0.01 \\
+ QLoRA Weights Average(S, T) & \textbf{26.24} & 27.14 $\pm$ 0.10 & \textbf{21.25 $\pm$ 0.04} & \textbf{30.92 $\pm$ 0.04} & 0.59 $\pm$ 0.15 & 0.52 $\pm$ 0.01 & 0.52 $\pm$ 0.01 & \textbf{0.70 $\pm$ 0.01} & \textbf{0.40 $\pm$ 0.03} & \cellcolor{gray!25}0.56 $\pm$ 0.05 \\
\hline
\end{tabular}
\caption{Control Effectiveness for \textbf{Multi-attribute Control} alongside single-attribute control results for comparison. All values are averages over 3 runs and standard deviation is reported for single dataset results. S=the Combined Sentiment dataset, T=the Combined Topic dataset, Ind mod=composition is on all 5 individually trained modules. Bold (shaded) =  (second) highest score in column and section; underline = train and test set from same dataset.
} 
\label{tab:multi_res_std_ce}
\end{table*}

\section{Hyperparameters}
\label{app:hyperparam}

\subsection{Trained Models}

We train all our QLoRA modules with the following pre-trained raw models: \texttt{meta-llama/Meta-Llama-3-8B}, \texttt{meta-llama/Llama-3.1-8B}, and \texttt{mistralai/Mistral-7B-v0.3}. Each model is trained on 3 different seeds (8989, 79817, 794323). All the executions are done on a NVIDIA A100 with 80GB.

We perform a small grid search for each trained module investigating the learning rate (5e-6, 2e-4) and learning rate scheduler (cosine, constant). We use Paged AdamW~\citep{loshchilov2017decoupled} as the optimiser. 

We set all the other hyperparameters as \textit{optimizer}=paged\_adamw\_32bit, \textit{batch size}=4, \textit{gradient accumulation steps}=4, \textit{maximum gradient norm}=1.0, \textit{warmup ratio}=0.1, \textit{weight decay}=0.5, \textit{group by length}=true, \textit{fp16}=false, \textit{bf16}=true, \textit{maximum sequence length}=1024, \textit{padding}=right, \textit{save strategy}=epoch and we trained the modules for up to 3 epochs. During inference we set \textit{batch size}=64.

Regarding model quantization, we set \textit{bnb 4bit compute dtype}=bfloat16, \textit{use 4bit}=True, \textit{bnb 4bit quant type}=nf4, and \textit{use nested quant}=False.

Regarding QLoRA hyperparameters, we follow the hyperparameters setting used in the vanilla QLoRA by \citeauthor{dettmers2023qlora} (Appendix~B.2). We set \textit{rank}=64, \textit{alpha}=16, \textit{dropout}=0.1 and we attach QLoRA at every linear layer of the pre-trained model (["q\_proj", "k\_proj", "v\_proj", "o\_proj", "gate\_proj", "up\_proj", "down\_proj", "lm\_head"]).

To get the best trained module for each run, we evaluate Control Effectiveness of each saved checkpoint using the validation portion of the train dataset. We determine the best model for each grid search based on the run with highest Control Effectiveness. Tables~\ref{tab:sent_best_mod} and~\ref{tab:top_best_mod} show the hyperamaters of the best-performing module for each executed grid search.

\begin{table}[]
    \centering
    \small
    \setlength\tabcolsep{2pt} % default value: 6pt
    \renewcommand{\arraystretch}{1.10}
    \begin{tabular}{|l|l|p{0.27\textwidth}|}
    \hline
    \textbf{Model} & \textbf{Seed} & \textbf{Hyperparameters} \\
    \hline
    \multirow{3}{*}{LLaMa 3 8B} & 8989 & \textit{checkpoint}=4562, \textit{learning rate scheduler}=cosine, \textit{learning rate}=5e-6 \\ \cline{2-3}
     & 79817 & \textit{checkpoint}=9124, \textit{learning rate scheduler}=cosine, \textit{learning rate}=5e-6 \\ \cline{2-3}
     & 794323 & \textit{checkpoint}=13686, \textit{learning rate scheduler}=constant, \textit{learning rate}=5e-6 \\
    \hline
    \multirow{7}{*}{LLaMa 3.1 8B} & 8989 & \textit{checkpoint}=4562, \textit{learning rate scheduler}=cosine, \textit{learning rate}=5e-6 \\ \cline{2-3}
     & 79817 & \textit{checkpoint}=13686, \textit{learning rate scheduler}=cosine, \textit{learning rate}=2e-4 \\ \cline{2-3}
     & 794323 & \textit{checkpoint}=13686, \textit{learning rate scheduler}=constant, \textit{learning rate}=5e-6 \\
    \hline
    \multirow{3}{*}{Mistral 7B} & 8989 & \textit{checkpoint}=13686, \textit{learning rate scheduler}=constant, \textit{learning rate}=5e-6 \\ \cline{2-3}
     & 79817 & \textit{checkpoint}=9124, \textit{learning rate scheduler}=constant, \textit{learning rate}=5e-6 \\ \cline{2-3}
     & 794323 & \textit{checkpoint}=4562, \textit{learning rate scheduler}=cosine, \textit{learning rate}=5e-6 \\
    \hline
    \end{tabular}
    \caption{Hyperparameters of the best-performing modules for sentiment control, selected through grid search. The modules were trained on the combined datasets.}
    \label{tab:sent_best_mod}
\end{table}

\begin{table}[]
    \centering
    \small
    \setlength\tabcolsep{2pt} % default value: 6pt
    \renewcommand{\arraystretch}{1.10}
    \begin{tabular}{|l|l|p{0.27\textwidth}|}
    \hline
    \textbf{Model} & \textbf{Seed} & \textbf{Hyperparameters} \\
    \hline
    \multirow{3}{*}{LLaMa 3 8B} & 8989 & \textit{checkpoint}=27997, \textit{learning rate scheduler}=constant, \textit{learning rate}=5e-6 \\ \cline{2-3}
     & 79817 & \textit{checkpoint}=27997, \textit{learning rate scheduler}=constant, \textit{learning rate}=5e-6 \\ \cline{2-3}
     & 794323 & \textit{checkpoint}=27997, \textit{learning rate scheduler}=constant, \textit{learning rate}=5e-6 \\
    \hline
    \multirow{3}{*}{LLaMa 3.1 8B} & 8989 & \textit{checkpoint}=27997, \textit{learning rate scheduler}=constant, \textit{learning rate}=5e-6 \\ \cline{2-3}
     & 79817 & \textit{checkpoint}=13998, \textit{learning rate scheduler}=cosine, \textit{learning rate}=5e-6 \\ \cline{2-3}
     & 794323 & \textit{checkpoint}=41995, \textit{learning rate scheduler}=constant, \textit{learning rate}=2e-4 \\
    \hline
    \multirow{3}{*}{Mistral 7B} & 8989 & \textit{checkpoint}=13998, \textit{learning rate scheduler}=cosine, \textit{learning rate}=2e-4 \\ \cline{2-3}
     & 79817 & \textit{checkpoint}=13998, \textit{learning rate scheduler}=cosine, \textit{learning rate}=5e-6 \\ \cline{2-3}
     & 794323 & \textit{checkpoint}=41995, \textit{learning rate scheduler}=constant, \textit{learning rate}=2e-4 \\
    \hline
    \end{tabular}
    \caption{Hyperparameters of the best-performing modules for topic control, selected through grid search. The modules were trained on the combined datasets.}
    \label{tab:top_best_mod}
\end{table}

\subsection{Evaluation Metrics}\label{app:eval_metr}

\paragraph{SLOR} We used \texttt{gpt2-xl} and \texttt{bigscience/bloom-1b7} models from HuggingFace to compute sentence and unigram probabilities.

\paragraph{Control Effectiveness} For sentiment control, we used \texttt{distilbert/distilbert-base-uncased- finetuned-sst-2-english}\footnote{\url{https://huggingface.co/distilbert/distilbert-base-uncased-finetuned-sst-2-english}} and \texttt{michelecafa gna26/t5-base-finetuned-sst2-sentiment}\footnote{\url{https://huggingface.co/michelecafagna26/t5-base-finetuned-sst2-sentiment}} from HuggingFace, and DeBERTa fine-tuned on Yelp~\citep{gu-etal-2023-controllable}. For topic control, we used \texttt{textattack/distilbert-base-uncased-ag- news},\footnote{\url{https://huggingface.co/textattack/distilbert-base-uncased-ag-news}} and \texttt{fabriceyhc/bert-base-uncased-ag\_ news}\footnote{\url{https://huggingface.co/fabriceyhc/bert-base-uncased-ag_news}} from HuggingFace, and DeBERTa~\citep{gu-etal-2023-controllable}. For topic control, we further used a general-purpose intruction-tuned LLM as classifier, namely \texttt{CohereForAI/c4ai-command-r-plus-4bit}.\footnote{\url{https://huggingface.co/CohereForAI/c4ai-command-r-plus-4bit}}

\section{Scientific artifacts and licensing}
\label{app:license}
Mistral 7B v0.3 and PPLM prompts are licensed under the Apache-2.0 license. LLaMa 3 8B\footnote{\url{https://huggingface.co/meta-llama/Meta-Llama-3-8B/blob/main/LICENSE}} and LLaMa 3.1 8B\footnote{\url{https://huggingface.co/meta-llama/Llama-3.1-8B/blob/main/LICENSE}} are licensed under a commercial license. Yelp Reviews dataset\footnote{\url{https://s3-media3.fl.yelpcdn.com/assets/srv0/engineering_pages/bea5c1e92bf3/assets/vendor/yelp-dataset-agreement.pdf}} is licensed under a commercial license. The DBPedia ontology classification dataset is licensed under the terms of the Creative Commons Attribution-ShareAlike License and the GNU Free Documentation License. IMDB, SST-2, and AG News datasets do not specify a license. STS benchmark test provides a license for each included dataset.\footnote{\url{http://ixa2.si.ehu.eus/stswiki/index.php/STSbenchmark}} regarding the classifiers used in the evaluation, DistilBERT and T5 fine-tuned on SST-2 are licensed under the Apach -2.0 license. BERT fine-tuned on AG News is licensed under the Apache-2.0 license, while DeBERTa models fine-tuned by~\citet{gu-etal-2023-controllable} are licensed under the MIT license. Command R plus is licensed under the Creative Commons Attribution Non Commercial 4.0. The usage of all listed artifacts is consistent with their licenses.

\end{document}